\documentclass[10pt,twocolumn,letterpaper]{article}

\usepackage[pagenumbers]{iccv} 
\usepackage{makecell}
\usepackage{multirow} 
\definecolor{iccvblue}{rgb}{0.21,0.49,0.74}
\usepackage[pagebackref,breaklinks,colorlinks,allcolors=iccvblue]{hyperref}

\def\paperID{1759} 
\def\confName{ICCV}
\def\confYear{2025}

\title{DreamLayer: Simultaneous Multi-Layer Generation via Diffusion Model}

\author{
    \textbf{Junjia Huang}$^{1,2}$\thanks{Equal Contribution.}\quad \textbf{Pengxiang Yan}$^{2*}$\quad \textbf{Jinhang Cai}$^{2}$\quad \textbf{Jiyang Liu}$^{2}$\\
    \textbf{Zhao Wang}$^2$\quad  \textbf{Yitong Wang}$^{2}$\quad  \textbf{Xinglong Wu}$^{2}$\quad  \textbf{Guanbin Li}$^{1,3,4}$\thanks{Corresponding Author.}\\
    {$^1$Sun Yat-sen University, $^2$ByteDance Intelligent Creation}, $^3$Peng Cheng Laboratory  \\
    $^4$Guangdong Key Laboratory of Big Data Analysis and Processing \\
    {\href{https://ll3rd.github.io/DreamLayer/}{https://ll3rd.github.io/DreamLayer/}}
}

\begin{document}
\maketitle
\begin{abstract}
Text-driven image generation using diffusion models has recently gained significant attention. To enable more flexible image manipulation and editing, recent research has expanded from single image generation to transparent layer generation and multi-layer compositions. However, existing approaches often fail to provide a thorough exploration of multi-layer structures,  leading to inconsistent inter-layer interactions, such as occlusion relationships, spatial layout, and shadowing. In this paper, we introduce DreamLayer, a novel framework that enables coherent text-driven generation of multiple image layers, by explicitly modeling the relationship between transparent foreground and background layers. DreamLayer incorporates three key components, i.e.,  Context-Aware Cross-Attention (CACA) for global-local information exchange, Layer-Shared Self-Attention (LSSA)  for establishing robust inter-layer connections, and Information Retained Harmonization (IRH) for refining fusion details at the latent level.
By leveraging a coherent full-image context, DreamLayer builds inter-layer connections through attention mechanisms and applies a harmonization step to achieve seamless layer fusion. 
To facilitate research in multi-layer generation, we construct a high-quality, diverse multi-layer dataset including $400k$ samples. Extensive experiments and user studies demonstrate that DreamLayer generates more coherent and well-aligned layers, with broad applicability, including latent-space image editing and image-to-layer decomposition.
\end{abstract}    
\section{Introduction}
\label{sec:intro}

In recent years, text-to-image generation based on diffusion models~\cite{rombach2022high, saharia2022photorealistic, betker2023improving, podell2023sdxl, chen2023pixart, chen2024pixart, zhang2024treereward} has demonstrated impressive capabilities to create high-quality, detail-rich images from text prompts. However, most methods focus on generating a single, complete image, significantly limiting their potential in applications like content editing and graphic design, which rely heavily on layered compositions. Layered structures are particularly advantageous for images containing multiple objects, as they allow for more flexible and versatile editing and creative modifications. This paper investigates the application of diffusion models to generate coherent, multi-layered images through a simple text-driven process.

\begin{figure}[t]
  \centering
   \includegraphics[width=1\linewidth]{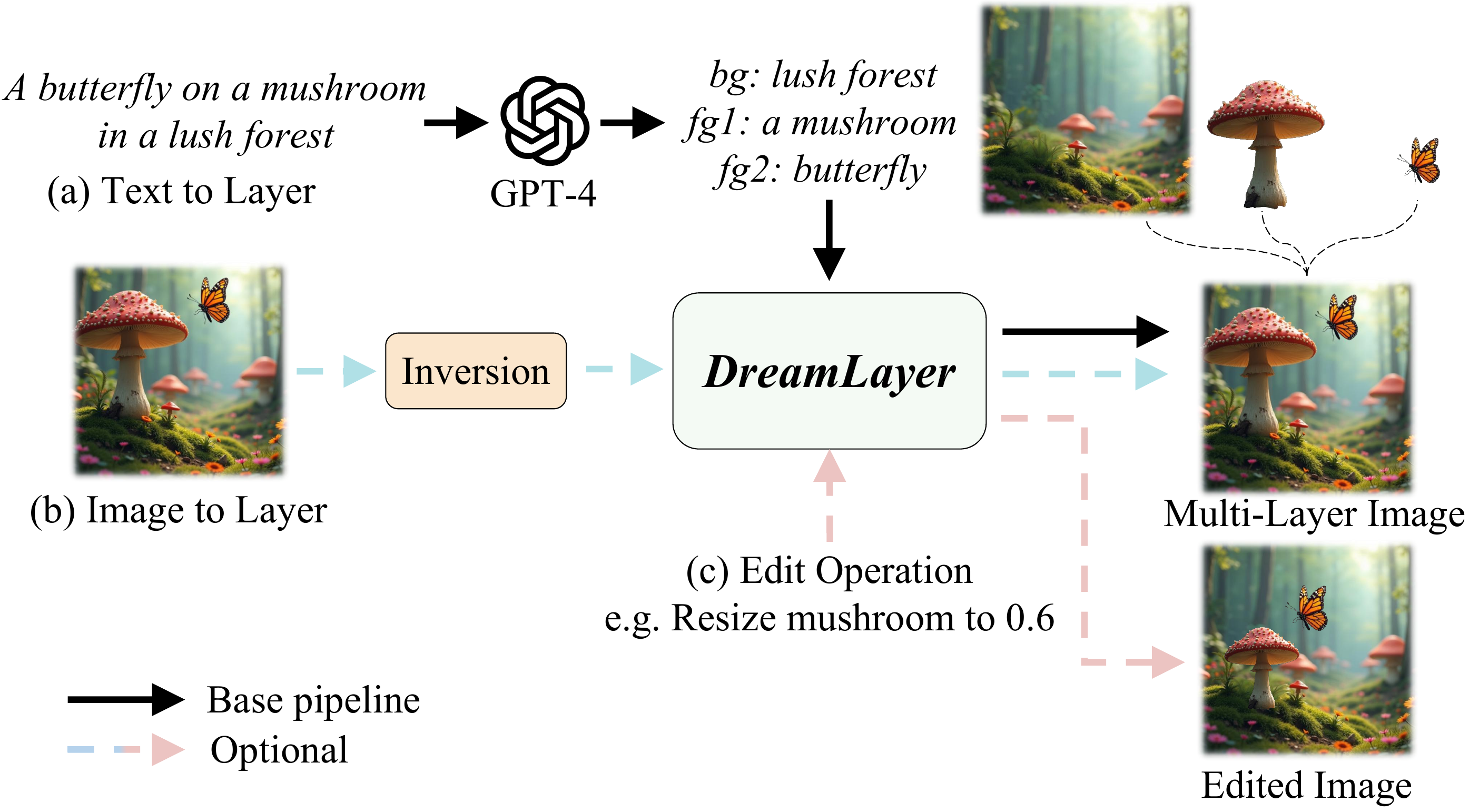}
   \caption{DreamLayer can handle multiple tasks: (a) Text-to-layer: Given a text input, we use GPT-4 to decompose foreground and background elements, feeding them into DreamLayer to generate a multi-layered image. (b) Image-to-layer: By using inversion to initialize starting latent, DreamLayer can decompose an image based on text prompts. (c) Latent-space editing: During denoising, DreamLayer can respond to editing instructions, producing more harmonious and consistent edited images.}
   \label{fig: intro_pipe}
\end{figure}

\begin{figure*}[t]
  \centering
   \includegraphics[width=1\linewidth]{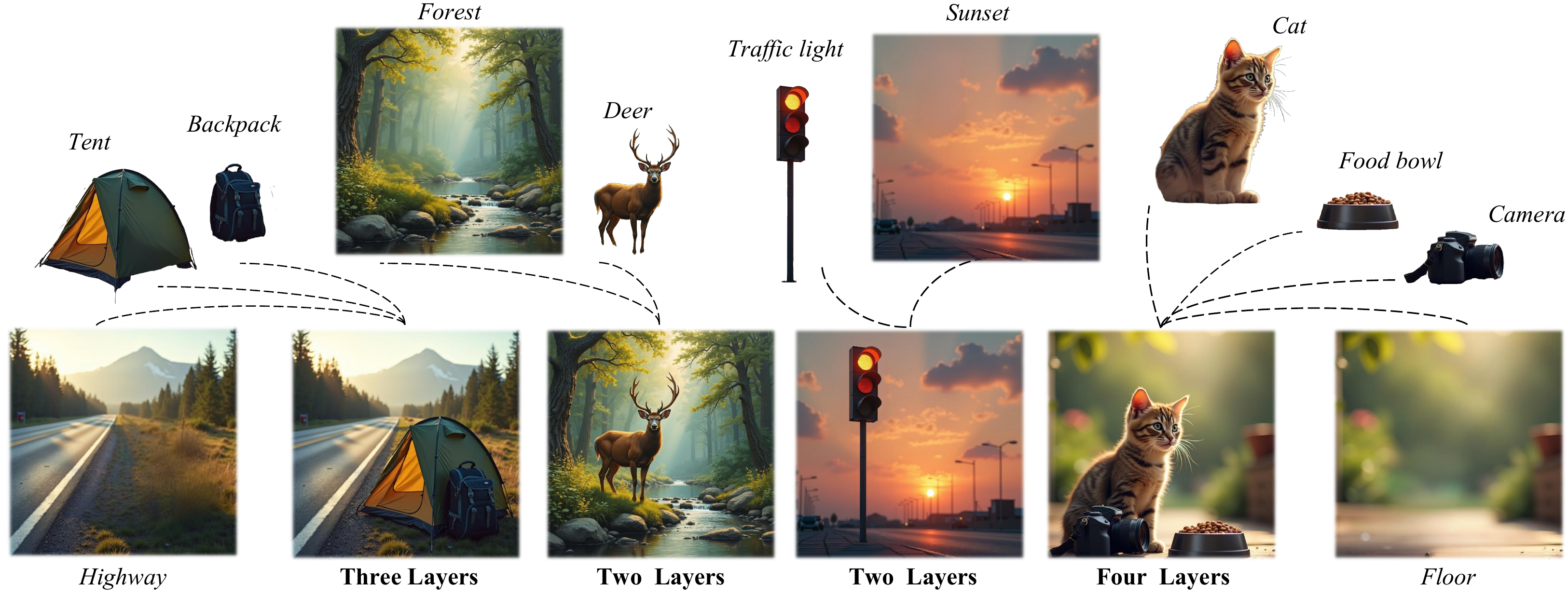}
   \caption{Multi-layer Dataset: Each image consists of a multi-layer structure, including a background and multiple foreground objects, with each foreground object represented as a transparent layer.}
   \label{fig: visual_intro}
\end{figure*}

Recent methods have started to explore the simultaneous generation of layered image structures to better support AI-driven image editing workflows. Most existing methods~\cite{zhang2023text2layer, zhang2024transparent} are limited to the generation of two-layer structures, \ie, foreground and background layers. While certain approaches~\cite{zhang2024transparent, huang2024layerdiff}  attempt to model the multi-layer generation task, they lack consideration for the relationship between different foreground layers and the background. For instance, LayerDiffusion~\cite{zhang2024transparent} disregards the spatial relationships between layers when adding new ones, leading to unintended overlaps between layers. LayerDiff~\cite{huang2024layerdiff} attempts to generate multi-layer composite images simultaneously, but it can only generate isolated, non-overlapping layers. These methods typically rely on simple stacking for layer composition, neglecting essential effects like shadows and occlusion, which are important for cohesive multi-layer generation and editing. Furthermore, a significant challenge in multi-layer generation is the lack of large-scale, high-quality open-source datasets. Current approaches and their datasets often rely on randomly stacked segmentation data~\cite{xu2024amodal}, suffer from limited data volume~\cite{tudosiu2024mulan}, or lack strictly defined multi-layered images~\cite{huang2024layerdiff}.


To address these challenges, we propose a comprehensive multi-layer data generation pipeline that decomposes images generated by advanced text-to-image models, createing a dataset of $400k$ multi-layer samples, as shown in \cref{fig: visual_intro}. 
In existing text-to-image generation models, when given a text prompt containing a background and multiple foregrounds, the models often demonstrate the ability to automatically arrange objects in a reasonable layout and generate harmonious compositions.
Building upon this, we introduce DreamLayer, a framework that utilizes global layer information to guide inter-layer attention and integrates a harmonization mechanism. To address layout issues in foreground layers, we begin by generating a cohesive global image from the full-text prompt. Subsequently, we employ Context-Aware Cross-Attention to extract contextual information from the global image, guiding the generation of foreground layers. To establish connections between layers, we adapt Layer-Shared Self-Attention, which further facilitates the sharing of global information across independent layers. 
Finally, we apply Information Retained Harmonization to fuse the composite image in latent space, ensuring a harmonious final result and improving flexibility and consistency for subsequent editing tasks.
DreamLayer enables the generation of multi-layer images with cohesive layouts and seamless integration across layers. It also supports a variety of tasks. As illustrated in \cref{fig: intro_pipe}, (a) DreamLayer can perform text-to-layer to generate multi-layer images by adaptively decomposing user text prompts using GPT-4; (b) DreamLayer supports image-to-layer decomposition by initializing the denoising latent via inversion and directing it based on text prompts in a training-free manner; (c) DreamLayer supports user-driven edits during the denoising process, ensuring stable and harmonious edited images. In summary, our key contributions are threefold:

\begin{itemize}
    \item We introduce DreamLayer, a simultaneous multi-layer generation framework that enhances harmony and consistency across layers via inter-layer interaction.
    \item We propose a layer-level harmonization approach to achieve smoother inter-layer blending, making it more adaptable for subsequent editing tasks.
    \item The release of a large-scale, high-quality multi-layer dataset, containing $400k$ meticulously curated multi-layer images, covering multiple objects and scenes.
\end{itemize}

\begin{figure*}[t]
  \centering
   \includegraphics[width=0.92\linewidth]{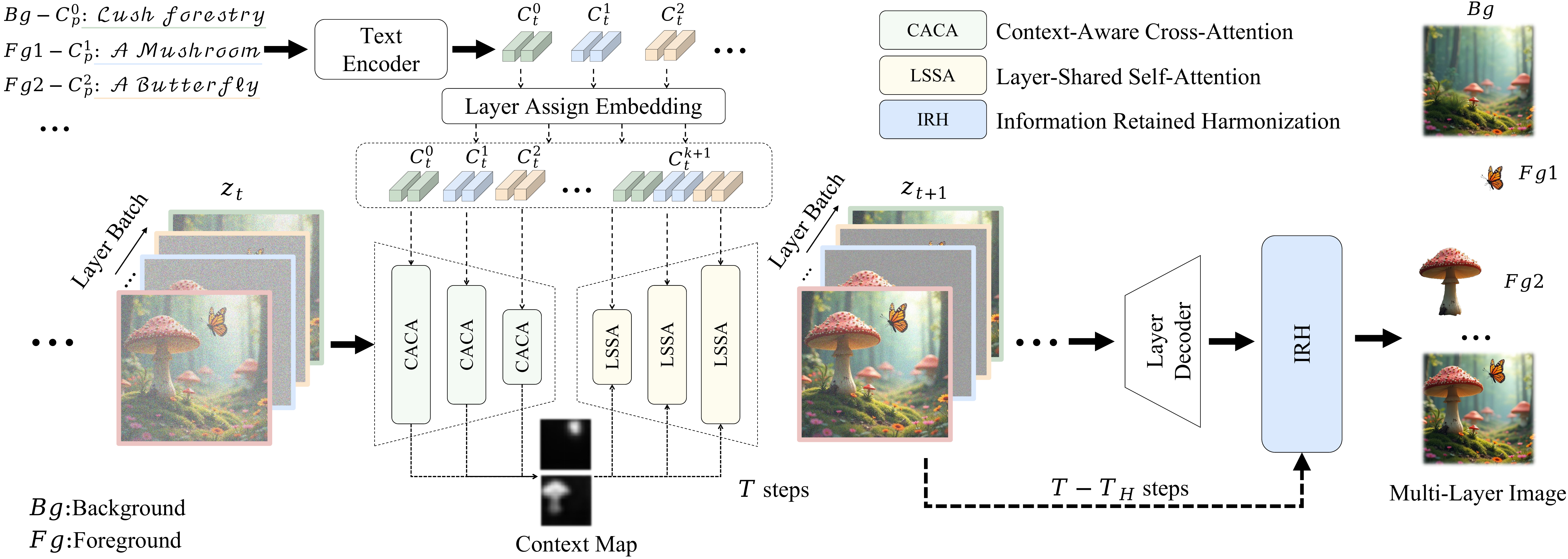}
   \caption{The DreamLayer Framework for Multi-Layer Image Generation: During the generation process, background and foreground prompts are combined via layer assign embeddings to form a global prompt $C_t^{k+1}$. In the attention phase, CACA extracts a context map from the global layer. Subsequently, the contextual information is fused across layers through LSSA, based on the global context map. Finally, IRH fuses the images using the latent image during the denoising process, achieving a harmonious result. }
   \label{fig: framework}
\end{figure*}

\begin{figure*}[t]
  \centering
   \includegraphics[width=0.92\linewidth]{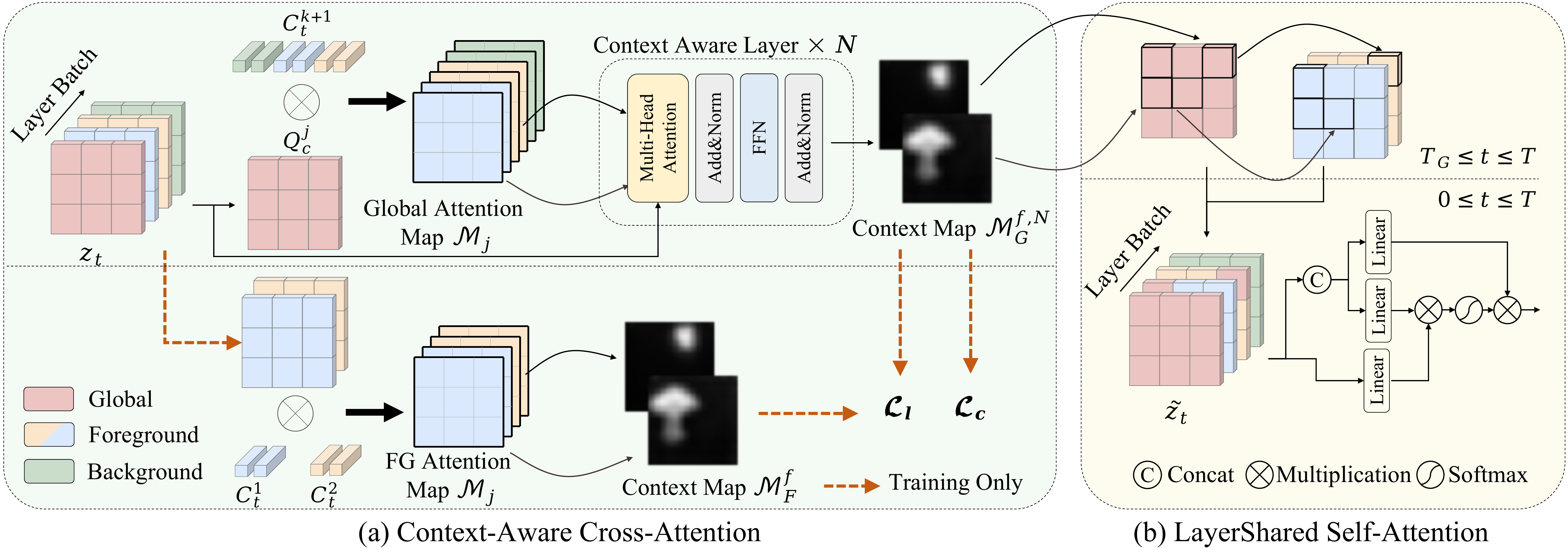}
   \caption{Overview of the Attention Mechanism in DreamLayer: (a) Context-Aware Cross-Attention for extracting the global context map and guiding the foreground layer layout; (b) Layer-Shared Self-Attention for establishing inter-layer connections and ensuring consistency.}
   \label{fig: framework_2}
\end{figure*}

\section{Related Work}
\textbf{Diffusion based Image Generation.} Diffusion models~\cite{ho2020denoising, song2020denoising} have shown leading performance in generative tasks, including image generation~\cite{zhu2023conditional, rombach2022high, zhang2024unifl}, editing~\cite{brooks2023instructpix2pix, huberman2024edit}, inpainting~\cite{liu2024structure}, and video generation~\cite{zhou2024storydiffusion}. These models have evolved from early pixel-space denoising~\cite{san2021noise} to latent-space denoising~\cite{rombach2022high}, with architectures progressing from U-Net~\cite{ronneberger2015u} to advanced designs like DiT~\cite{peebles2023scalable, chen2023pixart}. For multi-layer image generation, Text2Layer~\cite{zhang2023text2layer} uses a latent diffusion model to jointly denoise and reconstruct the RGB and alpha channels for two-layer images. LayerDiffusion~\cite{zhang2024transparent} encodes the alpha channel in the latent manifold and shares attention between foreground and background layers to generate multi-layered images. LayerDiff~\cite{huang2024layerdiff} proposes to generate multi-layer composites with layer-collaborative attention. However, these approaches often neglect integrated effects such as shadows and other inter-layer interactions between multiple foreground and background layers.

\noindent\textbf{Controllable Diffusion Model and Image Editing.} To enhance controllability in image generation, a range of methods have been developed. Textual Inversion~\cite{gal2022image} and DreamBooth~\cite{ruiz2023dreambooth} enable personalized content generation from a small set of example images. ControlNet~\cite{zhang2023adding} and T2I-Adapter~\cite{ye2023ip, mou2024t2i} introduce conditional signals, using reference images as visual prompts for direct guidance. Other methods~\cite{wang2024instancediffusion, zhou2024migc++} utilize bounding boxes to control image layout, while P2P~\cite{hertz2022prompt} and PnP~\cite{tumanyan2023plug} condition attention layers to manage content and style. To enable more customized image content, some methods~\cite{han2023improving, mokady2023null, huberman2024edit} leverage inversion techniques, converting the input image into a noise latent representation,  which is then edited and generated in a controlled manner based on text prompts. DesignEdit~\cite{jia2024designedit} further segment the latent representations into multiple layers, allowing more flexible spatial editing. However, existing methods typically restrict layers to non-overlapping structures.
In this work, we utilize a harmonious global layer to guide the generation of layers and employ independent layers to achieve seamless fusion in the latent space, enabling more harmonious editing.

\section{Methodology}
\label{sec:method}

\textit{\textbf{Definitions.}} Intuitively, a $k$-layer image consists of a background layer $I^1$, $k-1$ foreground layers $\{I^i\}_{i=2}^k$ and a global layer $I^{k+1}$. Each layer comprises a three-channel color image $I_c \in \mathbb{R}^{H \times W \times 3}$ and an alpha channel $I_\alpha \in \mathbb{R}^{H \times W \times 1}$, where the alpha channel indicates the visibility of the pixels within the color image. Formally, the global layer image $I^{k+1}$ can be expressed as
\begin{equation}
    I^{k+1} = \sum_{i=1}^k (I_\alpha^i \cdot I_c^i \cdot \prod^k_{f=i+1}(1-I_{\alpha}^f)).
\end{equation}
Each layer is associated with a corresponding textual description as a text prompt $\{C_p^i\}_{i=1}^{k+1}$.

\noindent\textit{\textbf{Generation of Alpha Channel.}} For each layer image, we fill the image with a solid gray background based on its alpha channel to obtain an RGB layer. This RGB layer is then encoded into a latent image $z\in\mathbb{R}^{hw\times D}$ and perturbed with noise for $t$ timesteps to produce a noisy latent image $z_t$. With the timestep $t$ and a text prompt $C$ as conditions, the diffusion model trains a network $\epsilon_\theta$ to predict the noise added to the noisy latent image $z_t$ with
\begin{equation}
    \mathcal{L}_{noise} = \mathbb{E}_{z_t, t, C, \epsilon \sim \mathcal{N}(0,1)}\left[||\epsilon-\epsilon_\theta(z_t, t, C)||_2^2\right],
\end{equation}
where $\mathcal{L}_{noise}$ represents the learning objective of the diffusion model. After $T$ denoising steps, the latent image $z_0$ is decoded by a layer decoder to generate the final transparent layer image with alpha channel. The layer decoder can be diverse, some methods~\cite{zhang2023text2layer, huang2024layerdiff} train a 4-channel VAE decoder, while others~\cite{zhang2024transparent} utilize a VAE decoder combined with a gray-background segmentation model. In this work, we adopt the same layer decoder as used in LayerDiffusion~\cite{zhang2024transparent}. Notably, we primarily focuses on the layout coherence and overall harmony in multi-layer generation, rather than the accuracy of alpha channel generation.

\noindent\textit{\textbf{Overview.}} As shown in \cref{fig: framework}, for multi-layer generation, we simultaneously encode the prompt of background layer and each foreground layer with the text encoder to obtain text embedding $\{C_t^i \in \mathbb{R}^{S\times D}\}_{i=1}^k$, where $S$ denotes the sequence length after tokenization. A learnable layer assign embedding is then added to each text embedding. We extract the portion between the [SOS] and [EOS] tokens from each text embedding and concatenate them to form a global embedding $C_t^{k+1}\in \mathbb{R}^{S\times D}$, which captures the essential information of all layers and guides the generation of the global layer. All layers are processed in a batch-wise manner during the attention computation of the diffusion model. To fully utilize the information from the global layer, we design three key components: Context-Aware Cross-Attention (CACA), Layer-Shared Self-Attention (LSSA), and Information Retained Harmonization (IRH). These components leverage guidance from the global layer to ensure consistency across background layer and foreground layers, facilitating the generation of harmonious multi-layer images.

\begin{figure*}[t]
  \centering
   \includegraphics[width=0.95\linewidth]{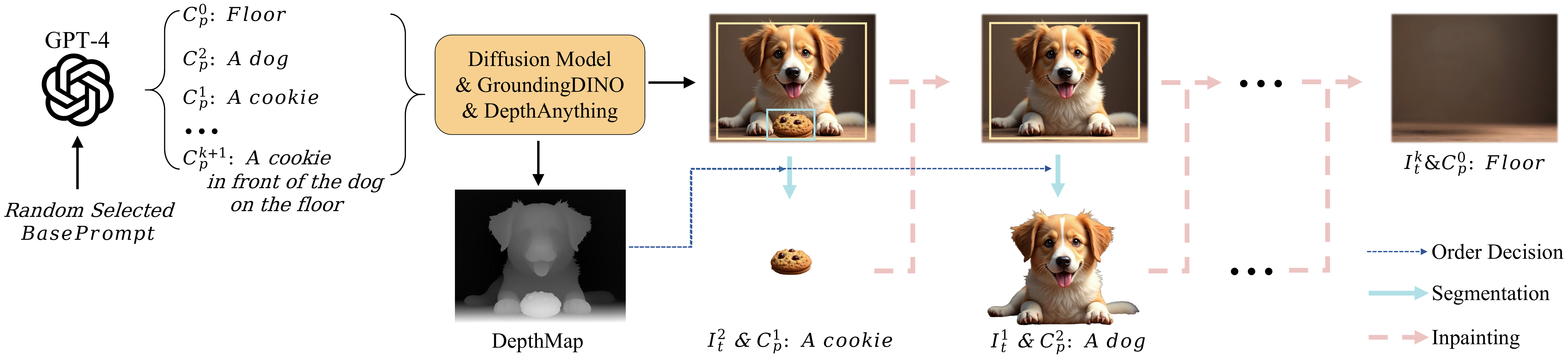}
   \caption{The pipeline of multi-layer data preparation. We utilize GPT-4 to process a randomly selected base prompt, structuring it into a background prompt and multiple foreground prompts. After generating the image using a diffusion model, we apply an open-set detection model GroundingDINO to identify the positions of the foreground objects and use the DepthAnything model to obtain a depth map. Based on the depth order, we sequentially extract the foreground layers and fill in the missing areas with an inpainting model.}
   \label{fig: data pipeline}
\end{figure*}

\subsection{Context-Aware Cross-Attention}
\label{sec: med_1}

The key to multi-layer generation is maintaining consistency in layout and proportions across all layers. During the generation process, we align the layout positions of other layers with the global layer. Utilizing the text embeddings of each layer, we extract relevant information from the cross-attention of the global layer.
Formally, as shown in \cref{fig: framework_2} (a), the global noisy latent image $z_t^{j,k+1}$ in the $j^{th}$ cross-attention mechanism is projected to a query matrix $Q_c^j=\ell_Q(z_t^{j,k+1})$ and the attention map $\mathcal{M}_j\in \mathbb{R}^{hw\times S}$ is then calculated with global embedding as 
\begin{equation}
    \mathcal{M}_j=Softmax(\frac{Q_c^j\ell_K(C_t^{k+1})^T}{\sqrt{d}}),
\end{equation}
where $\ell_Q, \ell_K$ are linear projections and $d$ is the latent dimension. The attention map preserves the spatial layout and geometry of the different foreground objects~\cite{hertz2022prompt, zhang2023diffusionengine}. Therefore, we extract the cross-attention maps corresponding to each foreground object from $J$ layers in the diffusion model. These maps are combined to create $f$ initial spatial-aware global attention maps $\mathcal{M}_{G}^f$:
\begin{equation}
    \mathcal{M}_{G}^f = Norm(\sum_{s=1}^{S_f}\sum_{j=1}^J(\mathcal{M}_j^s)), f=2,\cdots,k
\end{equation}
where $Norm(\cdot)$ denotes the Min-Max Normalization and $S_f$ denotes the token length of each foreground's text embedding within the global embedding.
To enhance foreground layer context in the extracted attention map, we feed the initial map and the global noisy latent image of $J$ cross-attention mechanism into $N$ context-aware layers $CAL(\cdot, \cdot)$ to generate global context map $\mathcal{M}_{G}^{f,n+1}$ as:
\begin{equation}
    \mathcal{M}_{G}^{f,n+1} = CAL(\mathcal{M}_G^{f,n}, \sum_{j=1}^J z_t^{j,k+1}).
\end{equation}
Each context-aware layer consists of a multi-head attention~\cite{vaswani2017attention} followed by a feed-forward network (FFN). The context map is supervised by the alpha channel of the foreground image with
\begin{equation}
    \mathcal{L}_c = \sum^f||\mathcal{R}(I^f_\alpha)-\mathcal{M}_{G}^{f,N}||_2,
\end{equation}
where $\mathcal{R}(\cdot)$ denotes the resize operation with interpolation.

After extracting the harmonious layout and geometric information of the foreground layer from the global layer $I^{k+1}$, we apply the same way to extract the corresponding spatial-aware attention maps $\mathcal{M}_F^f$ from the foreground-specific layers $(I^{i})_{i=2}^k$. Next, we implement a layout align loss $\mathcal{L}_{layout}$ to enable the global layer to supervise and guide the local foreground layers, facilitating alignment and coherence between them:
\begin{equation}
    \mathcal{L}_{layout} = \sum^f||\mathcal{M}_{G}^{f,N}-\mathcal{M}_F^f||_2.
\end{equation}
The final objective can be jointly written as
\begin{equation}
    \mathcal{L} = \lambda_{noise} \mathcal{L}_{noise} + \lambda_c \mathcal{L}_c + \lambda_{layout} \mathcal{L}_{layout},
\end{equation}
where $\lambda_{noise}, \lambda_c$ and $\lambda_{layout}$ are weight terms.

\subsection{Layer-Shared Self-Attention}
\label{sec: med_2}

To further strengthen the connections between layers, we propose a layer-shared self-attention. This approach first integrates global layer information into the foreground layers through the attention map, then processes information from all layers simultaneously within the self-attention mechanism, reinforcing inter-layer relationships and ensuring consistency throughout the multi-layer generation.

Specifically, as shown in \cref{fig: framework_2} (b), given a layer batch of noisy latent images $\{z_t^{i}\}_{i=1}^{k+1}$ and the global context map $\{\mathcal{M}_G^{i,t}\}^k_{i=2}$ at time step $t$, we integrate global information into the foreground layers based on the global context map, which is expressed as:
\begin{equation}
    \tilde{z}_t^i = z_t^{k+1} \cdot \mathcal{M}_G^{i,t} + z_t^i \cdot (1-\mathcal{M}_G^{i,t}).
    \label{eq10}
\end{equation}
For a diffusion model with $T$ denoising steps, we execute the process during the first $T_G$ steps.
Furthermore, to establish interaction between layers, we concatenate all noisy latent images along the sequence dimension to form a joint noise image at each denoising step:
\begin{equation}
    \tilde{z}_t^c = \mathrm{concat}(\tilde{z}_t^1, \cdots, \tilde{z}_t^{k+1}).
    \label{eq11}
\end{equation}
We then perform linear projections to generate the joint key $K_s^c$ and value $V_s^c$, and apply attention with original layer query $Q_s^i$, formally as:
\begin{equation}
    O_s^i = Softmax(\frac{Q_s^i(K_s^c)^T}{\sqrt{d}})V_s^c,
\end{equation}
where $d$ denotes the latent dimension. The weights of the linear projection are directly initialized from the original weights.

\subsection{Information Retained Harmonization}
\label{sec: med_3}
In multi-layer fusion, simply blending layers based on the alpha channel often affects the overall visual quality, as adding foreground objects to the background typically introduces shadow variations in real-world scenarios. To achieve a more harmonious fusion of the composite image, we propose Information Retained Harmonization, which blends latent during the denoising process and incorporates additional denoising steps, resulting in a more coherent and visually consistent final composite image.

\begin{table*}[t]
  \centering
  \footnotesize
  \begin{tabular}{cccccccccccc}
    \Xhline{1pt}
    \multirow{2}*{Methods} & \multicolumn{3}{c}{Two Layers} & & \multicolumn{3}{c}{Three Layers} & & \multicolumn{3}{c}{Four Layers} \\
    \cline{2-4} \cline{6-8} \cline{10-12}  
    \specialrule{0em}{2pt}{1pt}
    ~ & AES$\uparrow$ & Clip$\uparrow$ & FID$\downarrow$ & & AES$\uparrow$ & Clip$\uparrow$ & FID$\downarrow$ & & AES$\uparrow$ & Clip$\uparrow$ & FID$\downarrow$\\
    \Xhline{1pt}
    SD v1.5~\cite{rombach2022high} & 6.930 & 34.678 & 53.950 & & 6.363 & 34.222 & 55.198 && 6.367 & 35.000 & 59.149 \\
    LayerDiffusion~\cite{zhang2024transparent} & 6.522 & 32.466 & 63.481 && 6.058 & 30.350 & 67.118 && 5.975 & 29.158 & 79.997 \\
    DreamLayer w/o IRH & 6.967 & 34.587 & 51.957 & & 6.351 & 34.696 & 58.812 & & 6.340  & 34.993 & 57.073 \\
    DreamLayer & \textbf{7.013} & \textbf{34.835} & \textbf{50.761} & & \textbf{6.441} & \textbf{35.267} & \textbf{54.508} & & \textbf{6.422} & \textbf{35.723} & \textbf{53.598} \\
    \Xhline{1pt}
  \end{tabular}
  \caption{Quantitative comparison of multi-layer composite image generation. For SD v1.5, we generate the complete image from the global prompt as a composite image.}
  \label{tab:sota}
\end{table*}

Specifically, during the standard $T$ denoising steps, we retain the noisy latent images between step $T_H$ and $T_H'$, denoted as $\{z_t\}_{t=T_H}^{T_H'}$. After completing the $T$ denoising steps, we obtain the alpha channel $\{I_\alpha^i\}_{i=2}^k$ of the foreground-specific layer through the layer decoder. We then perform re-denoising for $T-T_H$ steps as a harmonization process, and between steps $T_H$ and $T_H'$, we conduct latent-level layer fusion. The formulation is as follows:
\begin{equation}
    \hat{z}_t^m = \hat{z}_t^1\cdot \prod_{i=2}^k(1-I_\alpha^i) + \sum_{i=2}^k(z_t^i\cdot I_\alpha^i \cdot \prod_{f=i+1}^k(1-I_\alpha^f)),
\end{equation}
where $\hat{z}$ represents the noisy latent image obtained during the harmonization steps. During the IRH process, the fused latent is influenced by the foreground objects throughout the denoising steps, allowing for the generation of corresponding shadow details and enhancing the overall coherence of the image. Simultaneously, information from the foreground layers is gradually preserved during denoising, ensuring the consistency of the generated foreground layers.

Additionally, we can edit the layers within the latent space, ensuring a smoother, more harmonious fusion of the layers. This is expressed as follows:
\begin{align}
    \hat{z}_t^m =& \hat{z}_t^1\cdot \prod_{i=2}^k(1-op(I_\alpha^i)) + \nonumber \\
    &\sum_{i=2}^k(op(z_t^i)\cdot op(I_\alpha^i) \cdot \prod_{f=i+1}^k(1-op(I_\alpha^f))),
    \label{eq:edit}
\end{align}
where $op(\cdot)$ represents the operations such as resizing, flipping, and moving. 

\begin{table}[t]
  \centering
  \footnotesize
  \begin{tabular}{lccccccccccc}
    \Xhline{1pt}
    Dataset & Images & Resolutions & Classes & Instances \\
    \Xhline{1pt}
    MuLAn~\cite{tudosiu2024mulan} & 44,860 & 600$\sim$800 & 759 & 101,269 \\
    DreamLayer & 408,187 & 896$\sim$1152 & 1453 & 525,388 \\
      \quad-TwoLayer & 305,801 & 896$\sim$1152 & 1379 & 305,801 \\
      \quad-ThreeLayer & 87,571 & 896$\sim$1152 & 1322 & 175,142 \\
      \quad-FourLayer & 14,815 & 896$\sim$1152 & 1045 & 44,445 \\
    \Xhline{1pt}
  \end{tabular}
  \caption{Dataset comparison between MuLAn and DreamLayer.}
  \label{tab_supp:mulan}
\end{table}

\subsection{Dataset Preparation}
\label{sec: med_4}

\cref{fig: data pipeline} illustrates the construction process of our multi-layer dataset. To manage complex layer relationships, we begin with the global layer and employ open-set object detection, depth maps, and inpainting to decompose it into multiple layers. 
First, we randomly sample a prompt from a large-scale prompt dataset~\cite{wang2022diffusiondb} as the base prompt. This base prompt it then processed by the GPT-4 model, which breaks it down into a background prompt $C_p^0$, several foreground prompts, and a complete global prompt $C_p^{k+1}$. If the base prompt lacks sufficient foreground objects, GPT-4 selects a suitable category from the Object365~\cite{shao2019objects365} dataset. Next, the global prompt is passed through a powerful image generation diffusion model, such as Flux~\cite{blackforest2024flux}, SD3~\cite{sd3}, or SDXL~\cite{podell2023sdxl}, to generate a complete image. We then use the foreground prompts and the open-set detection model, GroundingDINO~\cite{liu2023grounding}, to match the text with objects in the image. Simultaneously, we generate a depth map of the complete image using the DepthAnything~\cite{yang2024depth} model. Based on the depth map, we extract the object at the forefront using a matting model and fill in the missing areas with an inpainting model. Repeating this process, we determine the sequence of layers using the depth map and extract the corresponding foreground layers. We match the objects and text using segmentation masks and detection boxes, ultimately obtaining transparent images for multi-layers. Current generative models still struggle in generating a larger number of objects, resulting in low data retention. Therefore, we set the final output to 4 layers. Details of the pipeline are in the supplementary materials.

Following this pipeline, we generate a dataset containing millions of multi-layer images. We then conducted a manual review to filter and select images that met specific criteria, such as clear and complete foregrounds, harmonious backgrounds free of artifacts, and other quality standards. The final dataset comprises $300k$ two-layer images, $85k$ three-layer images, and $15k$ four-layer images. As shown in \cref{tab_supp:mulan}, our dataset contains more samples and encompasses a broader range of classes compared to existing datasets.

\section{Experiments}

\begin{figure}[t]
  \centering
   \includegraphics[width=0.85\linewidth]{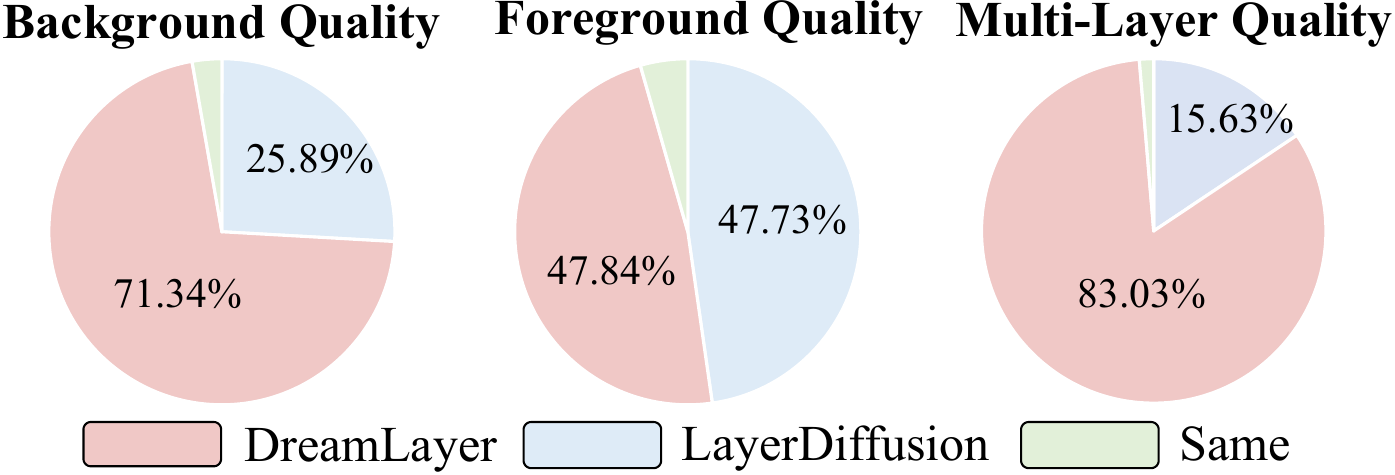}
   \caption{The vote preference percentage in user study. We evaluate our method and LayerDiffusion on three aspects: multi-layer, foreground, and background quality.}
   \label{fig: userstudy}
\end{figure}

\begin{figure*}[t]
  \centering
   \includegraphics[width=1\linewidth]{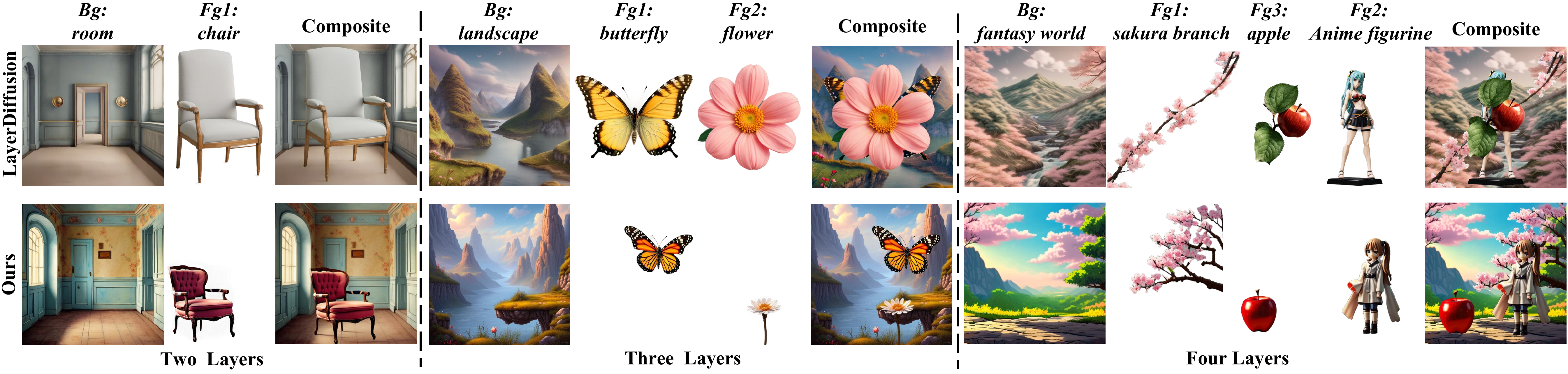}
   \caption{Qualitative comparison of multi-layer image generation. We present the generation results of Layerdiffusion and our method with two-layer, three-layer and four-layer images.}
   \label{fig: visual_layerdiffusion}
\end{figure*}

\subsection{Implementation Details}
\textbf{Training.} We initialize training with the pre-trained weights of Stable Diffusion v1.5~\cite{rombach2022high} and employ the Custom Diffusion~\cite{kumari2023multi} strategy, fine-tuning the K\&V linear layers in all attention layers. For foreground layers, additional K\&V layers are trained separately. The Context-Aware Cross-Attention is applied in the downsampling layers at a resolution of 16, while Layer-Shared Self-Attention is used in all upsampling layers. Each layer batch is initialized with same timestep noise, and the Layer Embedding is zero-initialized to minimize interference with the original weights. The training is performed over 4 days on 2 A100 GPUs with a batch size of 4 and a learning rate of 2e-6. More details are available in the supplementary materials.

\noindent\textbf{Evaluation.} We evaluate DreamLayer on a test set of 3k multi-layer images from our proposed dataset. Aesthetic quality is assessed using the AES Score~\cite{schuhmann2022laion}, text-image alignment with the CLIP-Score~\cite{radford2021learning}, and distribution similarity with the FID~\cite{heusel2017gans}.

\begin{figure}[t]
  \centering
   \includegraphics[width=0.9\linewidth]{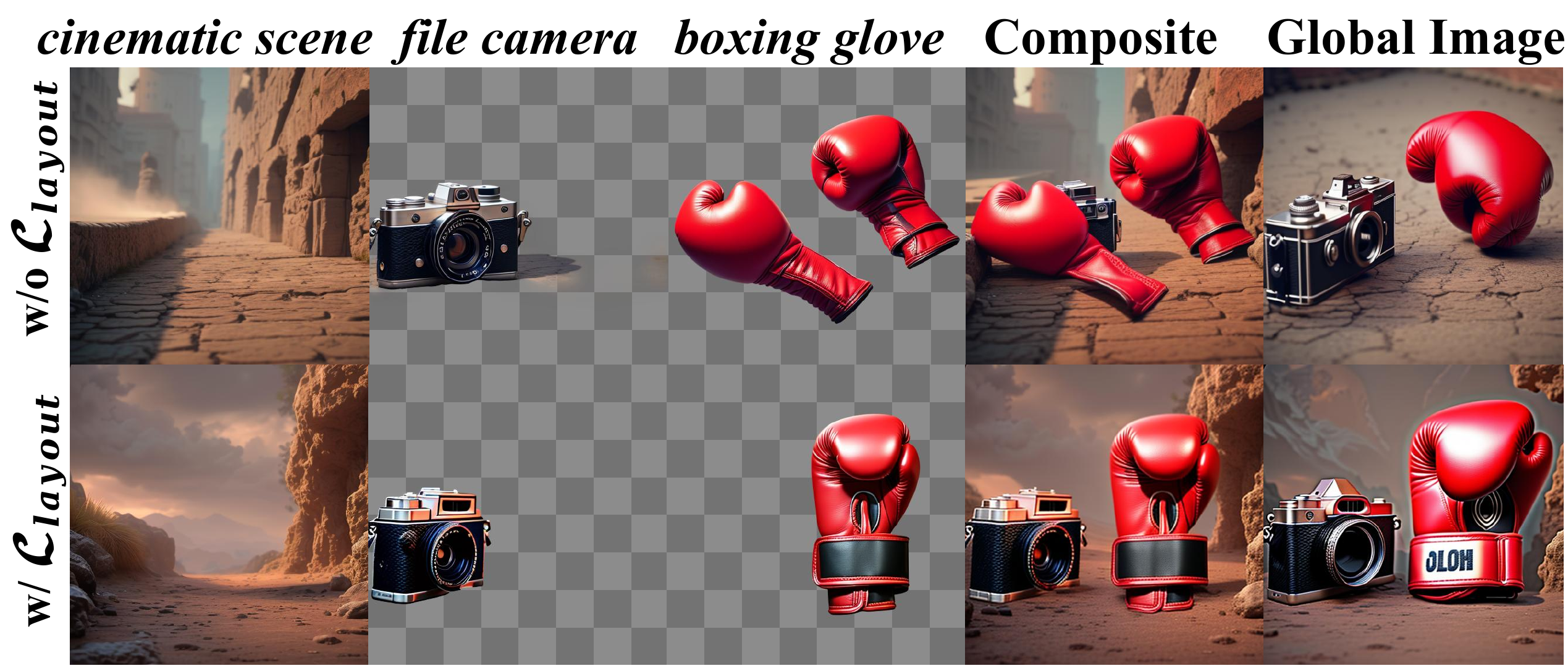}
   \caption{Ablation Study on Context-Aware Cross-Attention: Using $\mathcal{L}_{layout}$ supervision to extract layout information from the global image, guiding the generation of foreground layers and reducing overlapping placements.}
   \label{fig: ab_cross}
\end{figure}

\begin{table}[t]
  \centering
  \footnotesize
  \begin{tabular}{cccccccccccccc}
    \Xhline{1pt}
    \multicolumn{2}{c}{DreamLayer} & \multicolumn{3}{c}{Multi-Layers (Average)} \\
    \cline{3-5}
    \specialrule{0em}{2pt}{1pt}
    LSSA & CACA & AES$\uparrow$ & Clip$\uparrow$ & FID$\downarrow$ &\\
    \Xhline{1pt}
     ~ & ~ & 6.438 &	33.808 & 57.241\\
     \checkmark &  ~ & 6.471 & 33.727 & 56.598 \\
     ~ & \checkmark & 6.561 & 34.004 & 55.788 \\
    \checkmark & \checkmark & \textbf{6.625} & \textbf{35.275} & \textbf{52.956} \\
    \Xhline{1pt}
  \end{tabular}
  \caption{Ablation study on LSSA and CACA.}
  \label{tab:ab_re}
\end{table}

\subsection{Comparisons of Multi-Layer Image Generation}
\textbf{Quantitative Comparisons.} As shown in \cref{tab:sota}, We compare DreamLayer's performance in generating complete layers with results from Stable Diffusion~\cite{rombach2022high} (SD15) and LayerDiffusion~\cite{zhang2024transparent}. In this setup, SD15 generates a single complete image based solely on a global prompt. We use LayerDiffusion's background-to-foreground approach for three-layer and four-layer images, sequentially adding foreground elements to simulate multi-layer composition. As shown in the table, our method outperforms LayerDiffusion across all three metrics for multi-layer generation, with a notable improvement of around 0.5 in aesthetic score. For composite multi-layer images, our approach also achieves higher aesthetic quality and better text alignment compared to direct full-image generation by SD15.

\noindent\textbf{Qualitative Comparisons.} In \cref{fig: visual_layerdiffusion}, we present the multi-layer image generation results. Compared to Layerdiffusion~\cite{zhang2024transparent}, our method produces more coherent and appropriately sized foreground layers and achieves a more harmonious blending of the foreground and background.

\noindent\textbf{User Study.} As shown in \cref{fig: userstudy}, we perform a user study with 20 subjects on 200 samples to evaluate the multi-layer generation quality of our method and LayerDiffusion~\cite{zhang2024transparent} across three aspects: multi-layer, foreground, and background quality. The results show that our method achieves a preference percentage of 71.34\%, 47.84\%, 83.03\% w.r.t the above three aspects. It indicates our method delivers more cohensive layouts and higher quality, particularly in background and multi-layer images.

\subsection{Ablation Study}

\textbf{Context-Aware Cross-Attention.} CACA extracts the context map information from the global layer and utilizes $\mathcal{L}_{layout}$ to guide the layout of the foreground layer. As shown in \cref{fig: ab_cross}, without the layout alignment loss w/o $\mathcal{L}_l$, foreground objects tend to generate in the same position, leading to overlap and occlusion. We report the qualitative results in \cref{tab:ab_re}. Removing CACA significantly degrades image quality, reducing the overall AES score of the multi-layer generation by 0.154.

\noindent\textbf{Layer-Shared Self-Attention.} LSSA is primarily used to maintain consistency across different image layers. As shown in \cref{tab:ab_re}, the absence of LSSA leads to a significant drop in the CLIP score, decreasing by approximately 1.27.

\begin{figure}[t]
  \centering
   \includegraphics[width=0.9\linewidth]{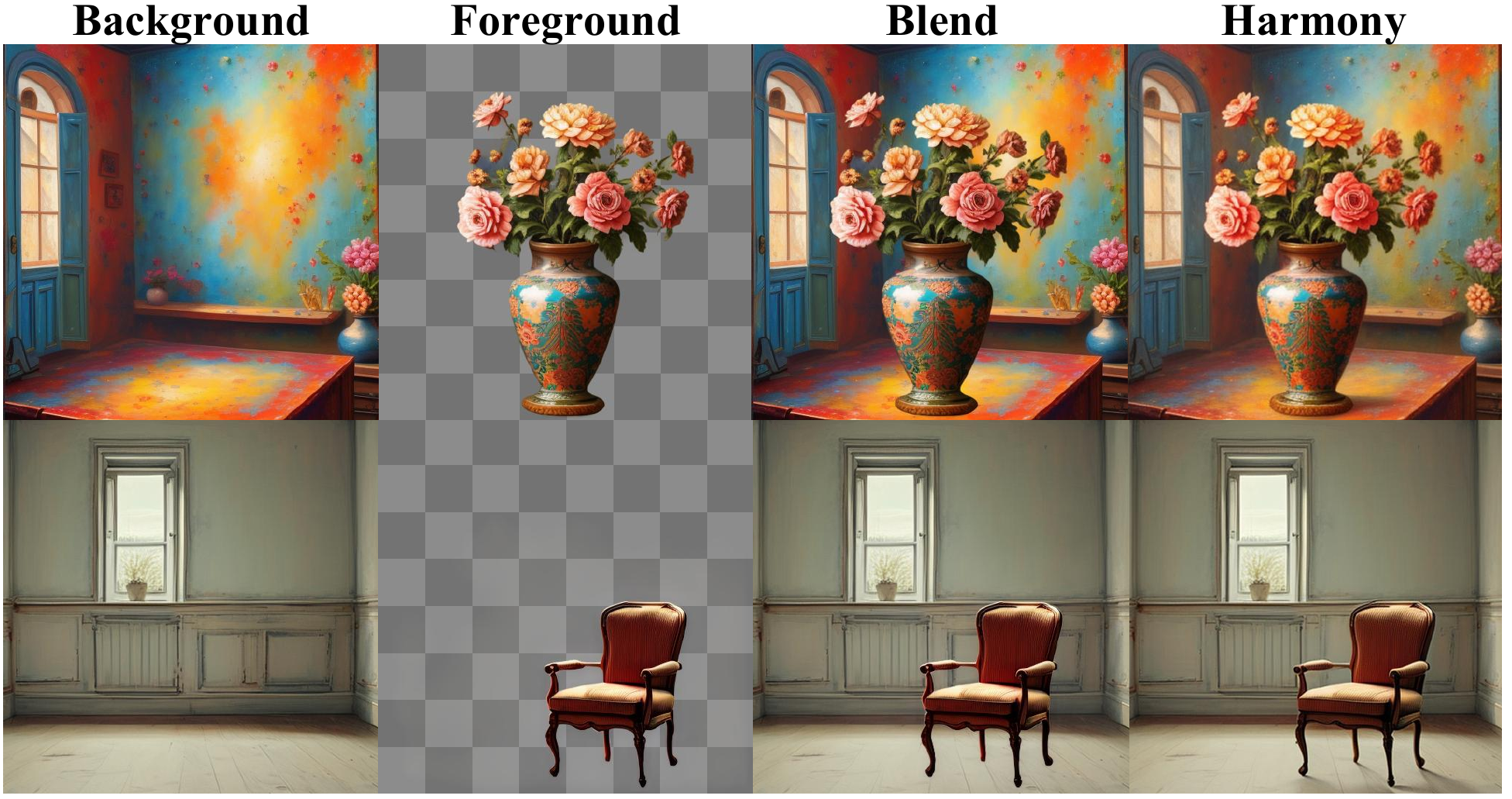}
   \caption{Ablation study on IRH. Our harmonization approach, unlike direct blending, generates appropriate shadows for foreground objects, resulting in a more cohesive overall composition.}
   \label{fig: ab_harm}
\end{figure}

\begin{table}[t]
  \centering
  \footnotesize
  \begin{tabular}{cccccccccccccc}
    \Xhline{1pt}
    $T_H$ & 0 & 200 & 400 & 600 & 800 \\
    $T_H'$ & 0 & 0 & 200 & 400 & 600 \\    
    \Xhline{1pt}
    \specialrule{0em}{2pt}{1pt}
     Avg AES$\uparrow$ & 6.553 & 6.568 & 6.600 & \textbf{6.625} & 6.615 \\
    \Xhline{1pt}
  \end{tabular}
  \caption{Investigation of $T_H$ and $T_H'$ in IBH.}
  \label{tab_supp: ibh}
\end{table}

\noindent\textbf{Information Retained Harmonization.} We further investigate the role of IRH in layer composition. As shown in \cref{fig: ab_harm}, simply stacking foreground and background layers (Blend) produces unrealistic composite images, lacking texture details like shadows. For example, the chair in \cref{fig: ab_harm} appears to float without a shadow, disrupting visual harmony. With IRH, however, shadows and other details are generated in the background to reflect the presence of foreground objects, resulting in a more natural and cohesive layer composition. For quantitative results, as shown in \cref{tab:sota}, ``DreamLayer w/o IRH" shows a noticeable decline in aesthetic scores, dropping by 0.1 without IRH. 

\noindent\textbf{$T_H$ and $T_H'$ in IBH.} We investigate the values of $T_H$ and $T_H'$ in the IBH. We experiment with $T_H$ from 800 to 0 steps. As \cref{tab_supp: ibh} shows, when $T_H' < 600$, IBH is applied near the end of the denoising process, resulting in poor harmonization and low AES score. Conversely, when $T_H$ is large (e.g., $T_H=800$), IBH over-modifies the background, reducing the AES score. Based on these observations, we selected $T_H=600$ and $T_H'=400$.

\subsection{Further Application}
\begin{figure}[t]
  \centering
   \includegraphics[width=0.9\linewidth]{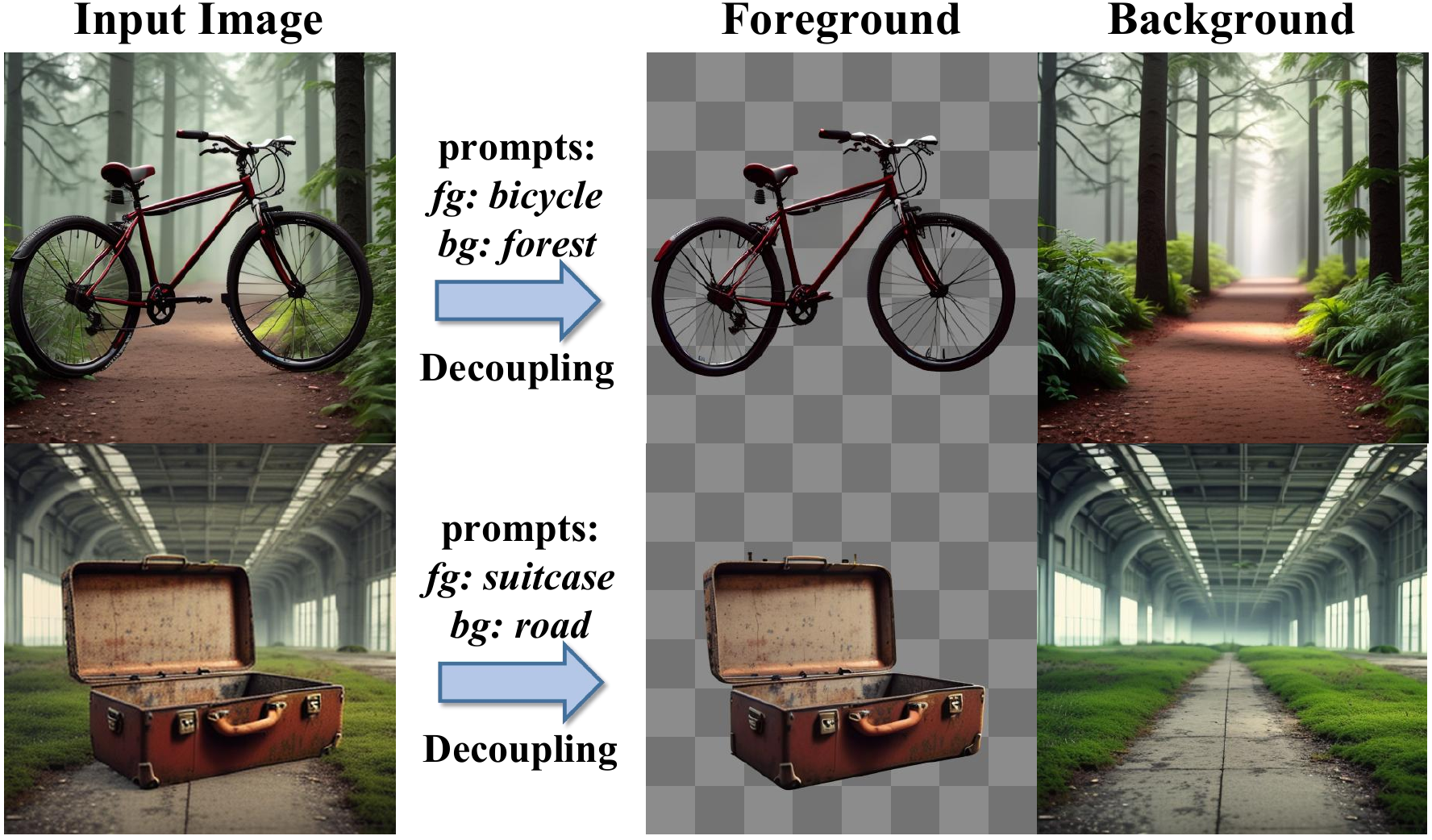}
   \caption{Image to Layer Visualization: By leveraging inversion to transfer the input image as the initial noise latent for all layers, DreamLayer can decompose the input  with the text prompt. }
   \label{fig: visual_img2layer}
\end{figure}

\noindent\textbf{Image to Layer}
Within the DreamLayer framework, we can extend it to Image-to-Layer task in a training-free manner. Specifically, we encode the input image into a latent representation as the global latent, then progressively add noise up to the $T$ step latent using an inversion technique~\cite{mokady2023null}, which serves as the initial latent for all layers in DreamLayer. To obtain a more accurate initial latent during this inversion process, we isolate global image information using a mask, minimizing the influence of other layers. As shown in \cref{fig: visual_img2layer}, this approach enables us to decompose the input image into separate layers based on text prompts. Detailed steps are provided in the supplementary materials.

\begin{figure}[t]
  \centering
   \includegraphics[width=0.9\linewidth]{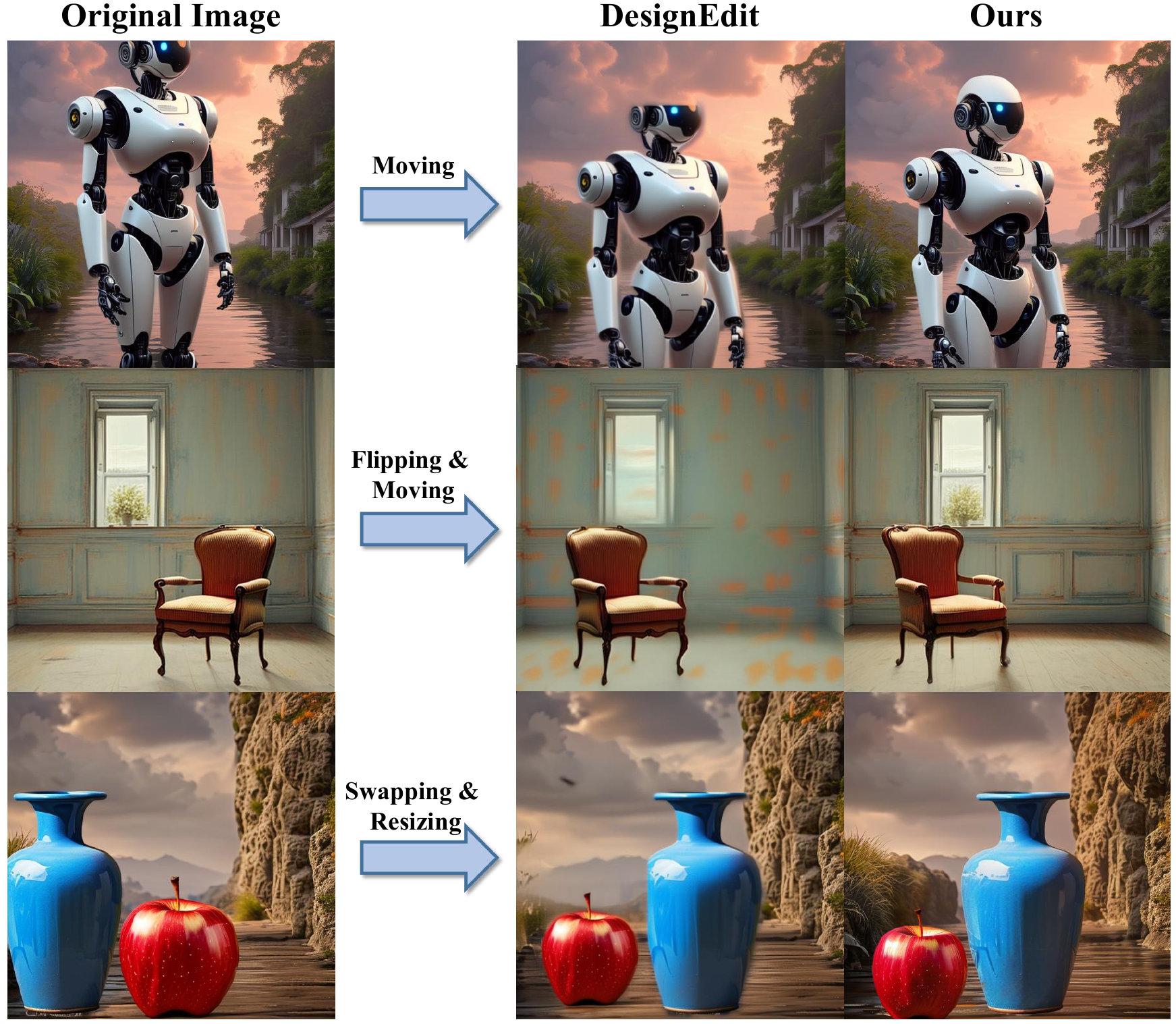}
   \caption{Layer Editing Visualization: Compared to DesignEdit, DreamLayer can complement objects at the image edges and create more cohesive results when they are flipped or moved.}
   \label{fig: visual_edit}
\end{figure}

\noindent\textbf{Layer Editing}
In practical applications, DreamLayer can generate multi-layer images and allow users to make harmonious edits to the layers. As described in \cref{eq:edit}, we perform these edits within IRH at the latent level, ensuring more cohesive adjustments. For instance, in \cref{fig: visual_edit}, when the chair is flipped and moved, the floor shadow is updated to align with its new position, enhancing overall realism.
Moreover, when parts of a foreground object extend beyond the image boundary, DreamLayer has the ability of foreground object amodal completion, which can complete the missing sections as needed when repositioned. As shown in \cref{fig: visual_edit}, compared to existing methods like DesignEdit~\cite{jia2024designedit}, DreamLayer successfully restores the out-of-frame areas of objects such as the robot and blue vase after they are moved.

\section{Conclusion}
In this paper, we introduce a large-scale, high-quality multi-layer dataset featuring diverse foreground objects and backgrounds. Building on this, we propose DreamLayer, a framework for simultaneously generating multi-layer images. To address layout consistency among foreground layers, we introduce Context-Aware Cross-Attention, which guides foreground generation using the harmonious layout of a global image. To enhance inter-layer connections, we present Layer-Shared Self-Attention, enabling effective information exchange between layers. Finally, to generate a cohesive composite image, we propose Information Retained Harmonization, which merges layers at the latent level to achieve seamless fusion. DreamLayer support not only multi-layer generation but also layer decomposition for image-to-layer task with inversion, enabling flexible editing within the latent space for harmonious adjustments. Experimental results demonstrate the effectiveness of DreamLayer in multi-layer generation.

{
    \small
    \bibliographystyle{ieeenat_fullname}
    \bibliography{main}
}
\clearpage
\appendix
\section{Multi-Layer Dataset}

\subsection{Pipeline of Data Generation.}
\begin{figure}[t]
  \centering
   \includegraphics[width=1\linewidth]{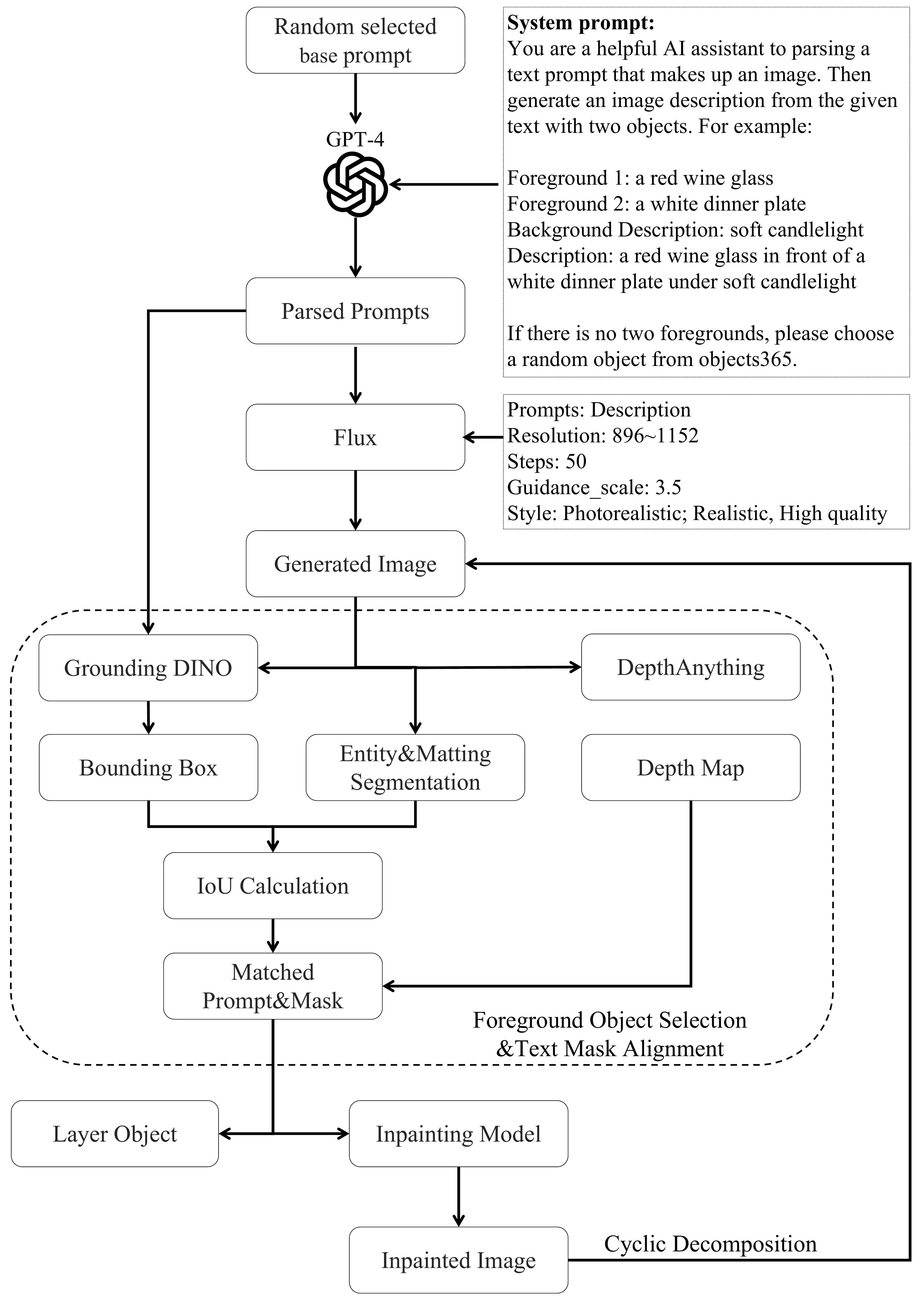}
   \caption{The pipeline of multi-layer data preparation. }
   \label{fig_supp: data pipeline}
\end{figure}

The detailed process for multi-layer data generation is illustrated in \cref{fig_supp: data pipeline}. First, a prompt is randomly selected from a large prompt dataset diffusiondb~\cite{wang2022diffusiondb}. Subsequently, this prompt is processed by GPT-4 to generate corresponding foregrounds, backgrounds, and a complete descriptive prompt. The descriptive prompt is fed into generation models like Flux to create images with resolutions ranging from 892 to 1152.
Next, GroundingDINO~\cite{liu2023grounding} and the foreground prompts are used to extract bounding boxes for the foreground objects from the generated image. Entity segmentation identifies all entities in the image. Based on the depth map~\cite{yang2024depth}, the foremost entity is selected. After matching it with the bounding box using IoU, the entity mask is linked to the text prompt.
We then refine the entity mask using a matting segmentation model, producing more detailed alpha channels and foreground layers. Finally, an inpainting model uses the foreground mask to fill in the image. This process is repeated to decompose all foregrounds and backgrounds, resulting in complete foreground and background layers.

Through this process, we automatically generated millions of multi-layer images. After manual filtering, we remove low-quality layers, such as those with foreign objects in the completed backgrounds, inaccurate foreground segmentation, or poor foreground quality. Finally, $400k$ high-quality layer data is retained. 
\subsection{Dataset Analysis}
\begin{figure*}[htbp]
    \centering
    \begin{subfigure}[t]{0.33\textwidth}
        \centering
        \includegraphics[width=\textwidth]{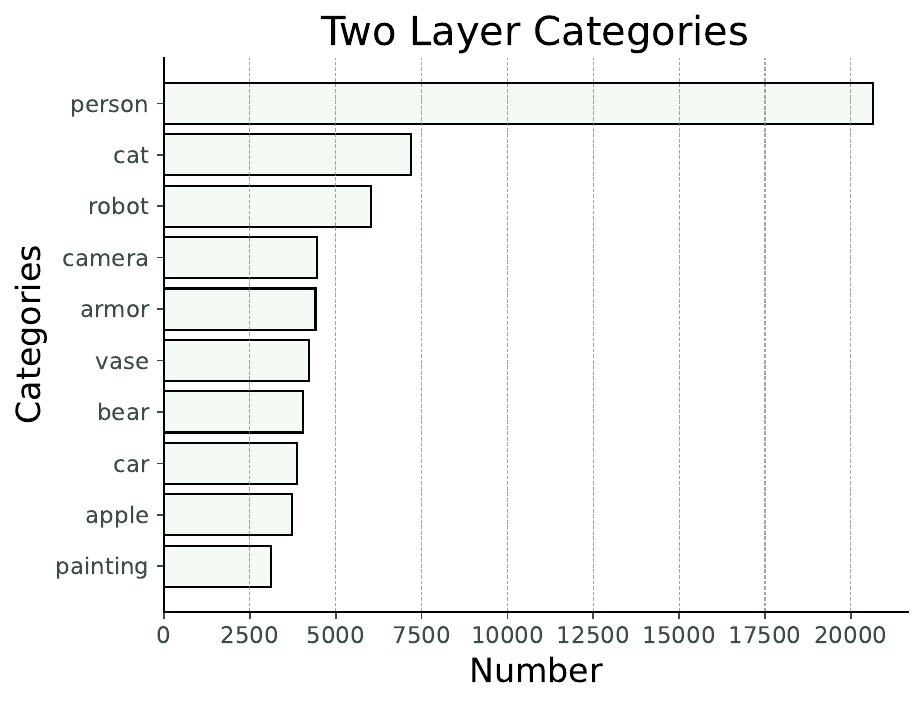}
    \end{subfigure}
    \hfill
    \begin{subfigure}[t]{0.33\textwidth}
        \centering
        \includegraphics[width=\textwidth]{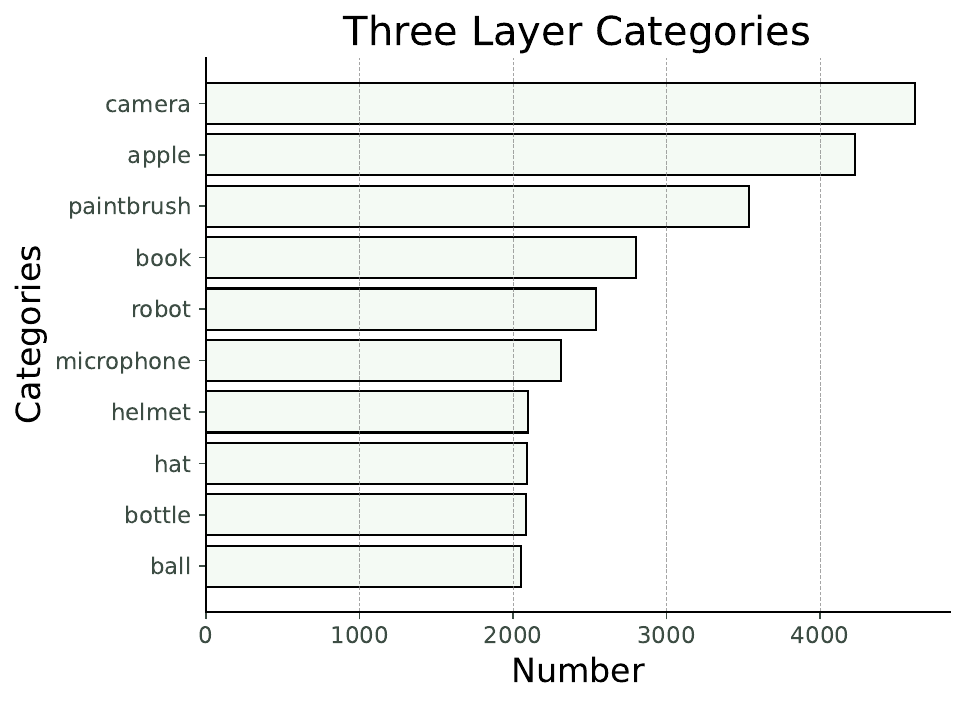}
    \end{subfigure}
    \hfill
    \begin{subfigure}[t]{0.33\textwidth}
        \centering
        \includegraphics[width=\textwidth]{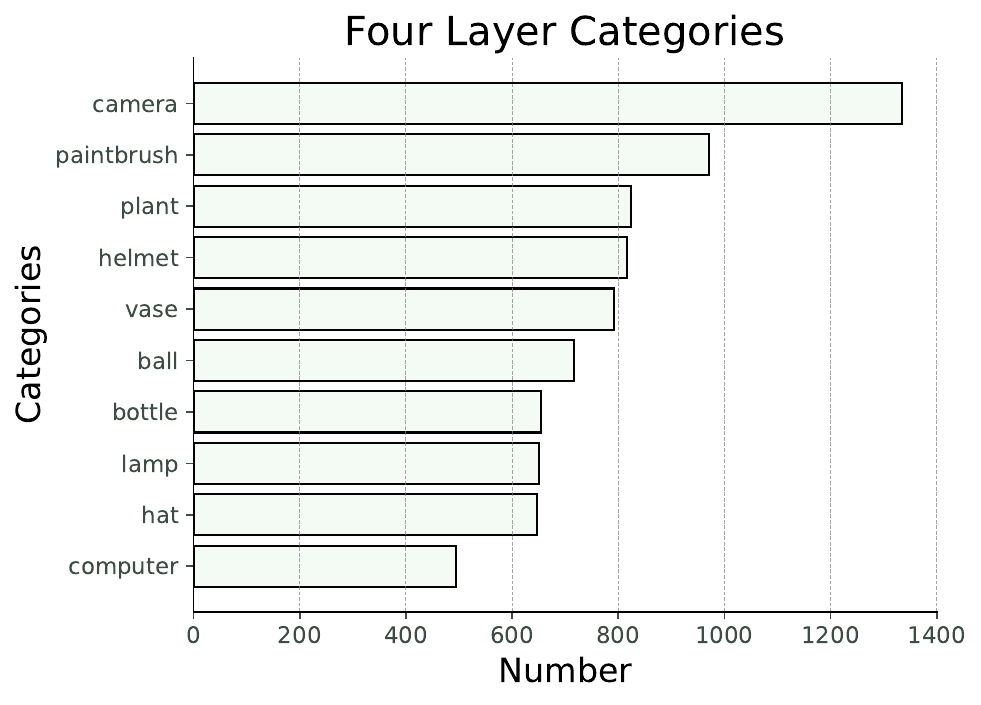}
    \end{subfigure}
    \caption{Top 10 most common categories in our Multi-Layer Dataset.}
    \label{fig_supp: categories}
\end{figure*}

We provide a  detailed comparison between our dataset and MuLAn~\cite{tudosiu2024mulan} in \cref{tab_supp:mulan}. Compared to the MuLAn dataset, we have more images, higher resolution, more categories, and a greater number of instances. 
\cref{fig_supp: categories} illustrates the top ten most common categories across multi-layer images. In the two-layer data, ``person'' is the dominant category, largely due to the abundance of portrait examples in the prompts. We deliberately reduced the generation of ``person'' instances in the three-layer and four-layer datasets, resulting in a more balanced category distribution for these layers.

\begin{table}[t]
  \centering
  \footnotesize
  \begin{tabular}{lccccccccccc}
    \Xhline{1pt}
    Dataset & Images & Resolutions & Classes & Instances \\
    \Xhline{1pt}
    MuLAn~\cite{tudosiu2024mulan} & 44,860 & 600$\sim$800 & 759 & 101,269 \\
    DreamLayer & 408,187 & 896$\sim$1152 & 1453 & 525,388 \\
      \quad-TwoLayer & 305,801 & 896$\sim$1152 & 1379 & 305,801 \\
      \quad-ThreeLayer & 87,571 & 896$\sim$1152 & 1322 & 175,142 \\
      \quad-FourLayer & 14,815 & 896$\sim$1152 & 1045 & 44,445 \\
    \Xhline{1pt}
  \end{tabular}
  \caption{Dataset comparison between MuLAn and DreamLayer.}
  \label{tab_supp:mulan}
\end{table}

\begin{table*}[t]
  \centering
  \small
  \begin{tabular}{cccccccccccc}
    \Xhline{1pt}
    \multirow{2}*{Methods (Bg)} & \multicolumn{3}{c}{Two Layers} & & \multicolumn{3}{c}{Three Layers} & & \multicolumn{3}{c}{Four Layers} \\
    \cline{2-4} \cline{6-8} \cline{10-12}  
    \specialrule{0em}{2pt}{1pt}
    ~ & AES$\uparrow$ & Clip$\uparrow$ & FID$\downarrow$ & & AES$\uparrow$ & Clip$\uparrow$ & FID$\downarrow$ & & AES$\uparrow$ & Clip$\uparrow$ & FID$\downarrow$\\
    \Xhline{1pt}
    LayerDiffusion~\cite{zhang2024transparent} & 6.034 & 28.426 & 81.491 & & 5.438 & 27.839 & 95.813 & & 5.564 & 28.907 & 117.485 \\
    DreamLayer & \textbf{6.731} & \textbf{29.827} & \textbf{72.633} & & \textbf{6.127} & \textbf{29.297} & \textbf{87.927} & & \textbf{6.119} & \textbf{30.661} & \textbf{80.157} \\
    \Xhline{1pt}
        \multirow{2}*{Methods (Fg)} & \multicolumn{3}{c}{Two Layers} & & \multicolumn{3}{c}{Three Layers} & & \multicolumn{3}{c}{Four Layers} \\
    \cline{2-4} \cline{6-8} \cline{10-12}  
    \specialrule{0em}{2pt}{1pt}
    ~ & AES$\uparrow$ & Clip$\uparrow$ & FID$\downarrow$ & & AES$\uparrow$ & Clip$\uparrow$ & FID$\downarrow$ & & AES$\uparrow$ & Clip$\uparrow$ & FID$\downarrow$\\
    \Xhline{1pt}
    LayerDiffusion~\cite{zhang2024transparent} & 6.124 & 30.404 & 64.406 & & 5.782 & 29.849 & 43.889 & & 5.652 & 29.646 & 45.210 \\
    DreamLayer & \textbf{6.165} & \textbf{30.530} & \textbf{51.495} & & \textbf{5.806} & \textbf{29.905} & \textbf{33.462} & & \textbf{5.703} & \textbf{29.724} & \textbf{31.426} \\
    \Xhline{1pt}
  \end{tabular}
  \caption{Quantitative comparison of background and foreground image generation.}
  \label{tab_supp:bg}
\end{table*}

\begin{figure}[t]
  \centering
   \includegraphics[width=0.9\linewidth]{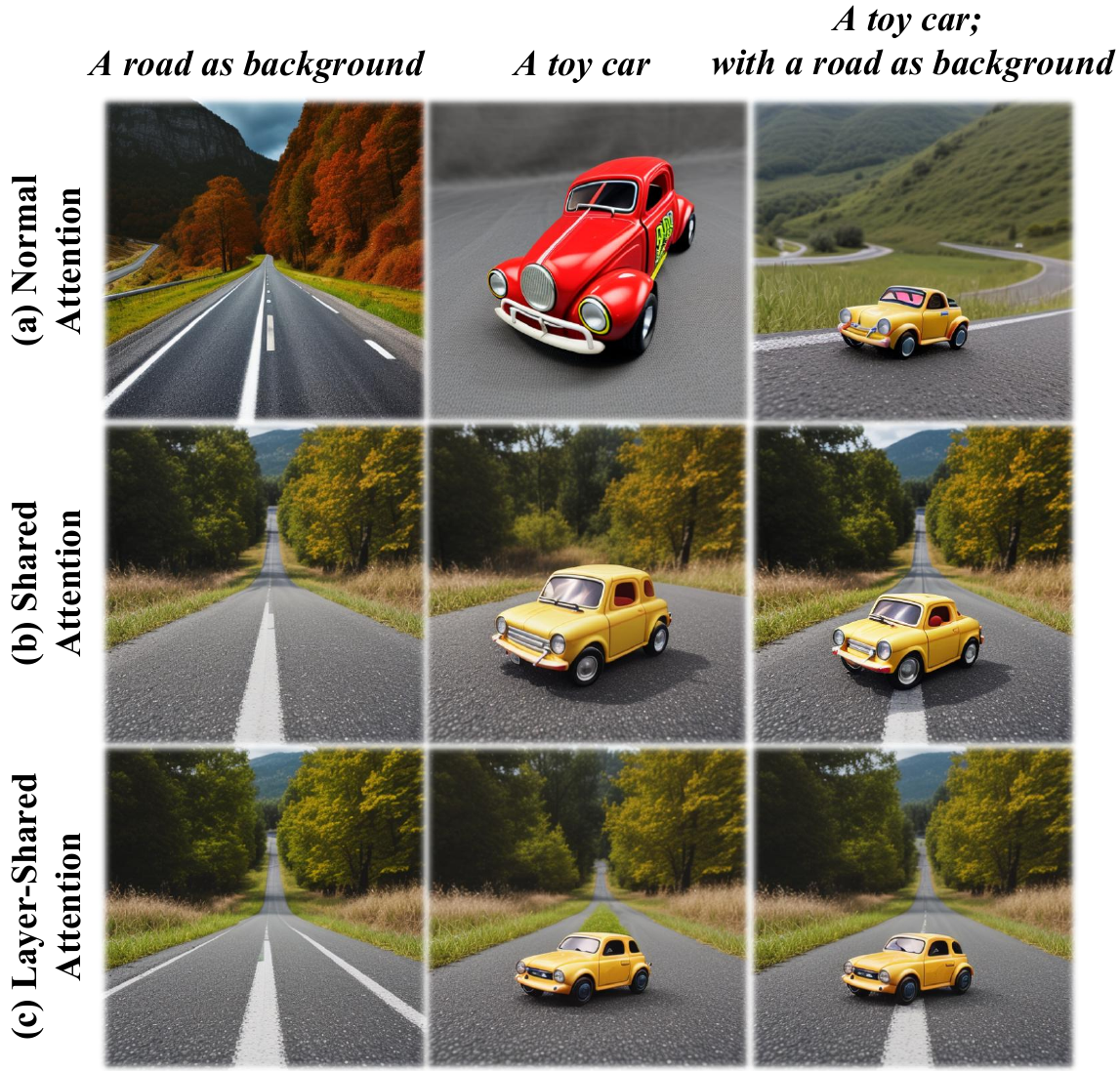}
   \caption{Ablation Study on Layer-Shared Attention: ``Normal Attention" refers to standard self-attention in SD15; ``Shared Attention" involves layer interaction through concatenation and ``Layer-Shared Attention" incorporating global layer information.}
   \label{fig: ab_self}
\end{figure}

\begin{figure}[t]
  \centering
   \includegraphics[width=1\linewidth]{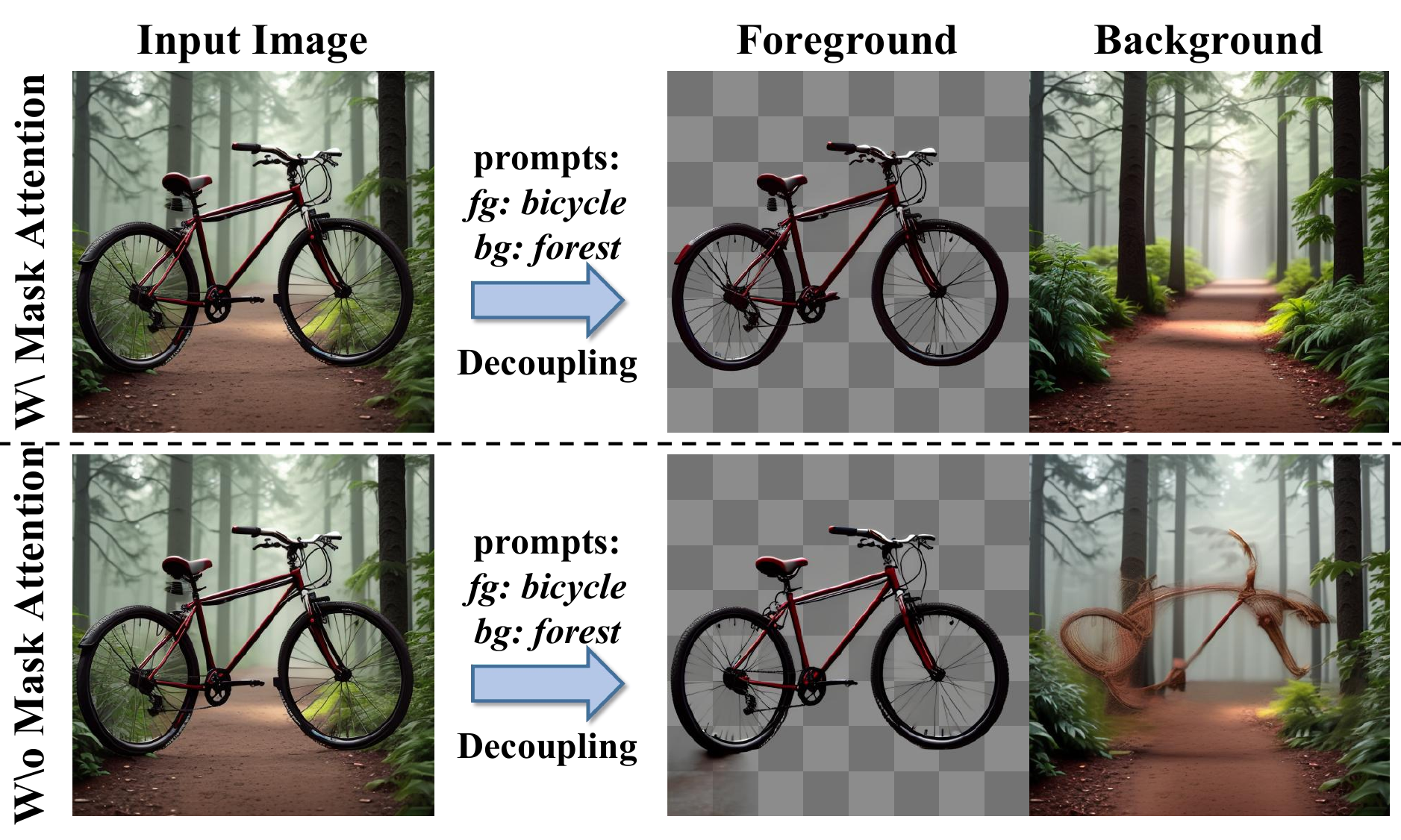}
   \caption{The effectiveness of mask attention in Image to Layer. }
   \label{fig_supp: image2layer}
\end{figure}

\subsection{Visualization}

\begin{figure*}[t]
  \centering
   \includegraphics[width=1\linewidth]{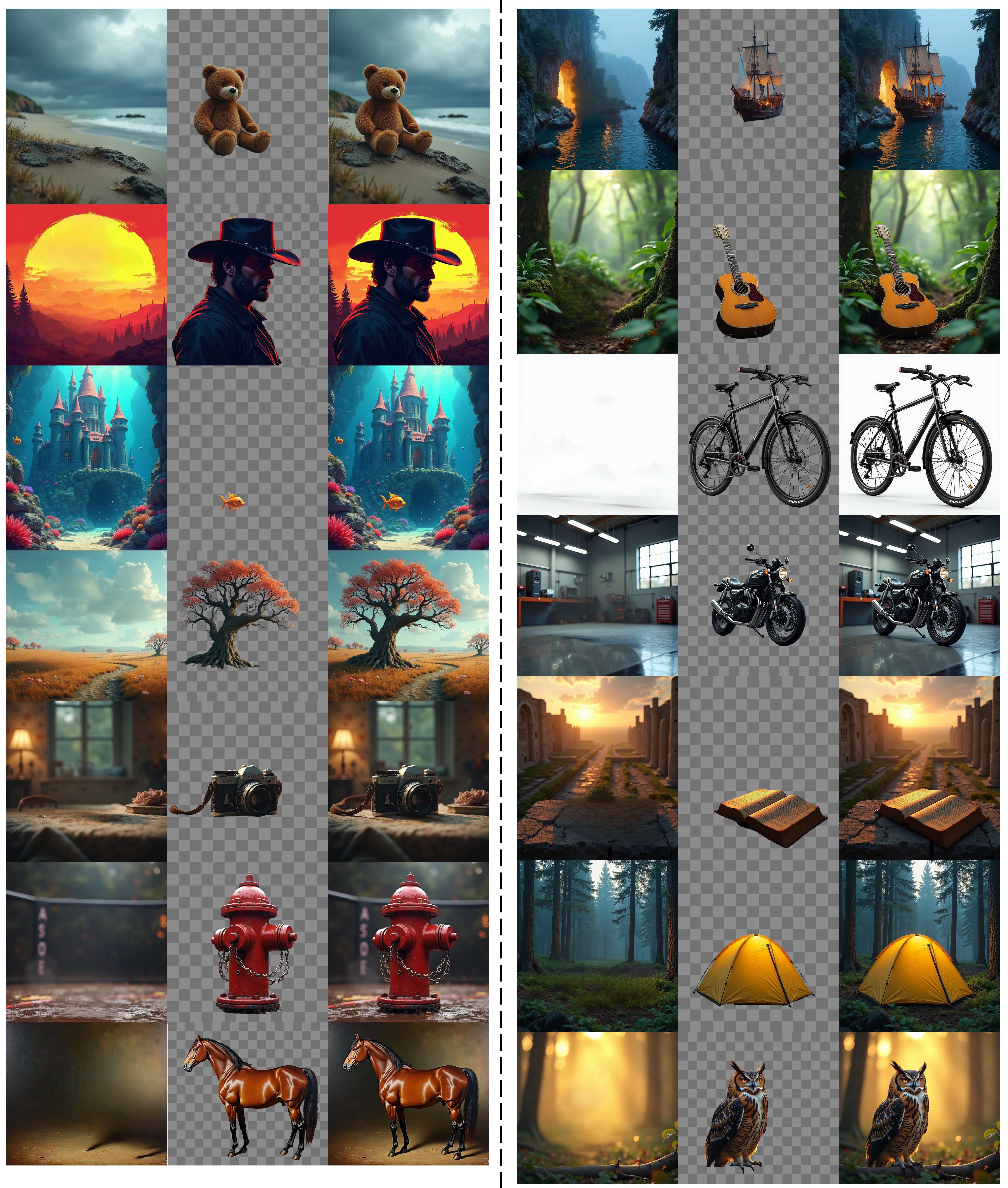}
   \caption{Visulization of two-layer images in our multi-layer dataset. }
   \label{fig_supp: two_dataset}
\end{figure*}

\begin{figure*}[t]
  \centering
   \includegraphics[width=1\linewidth]{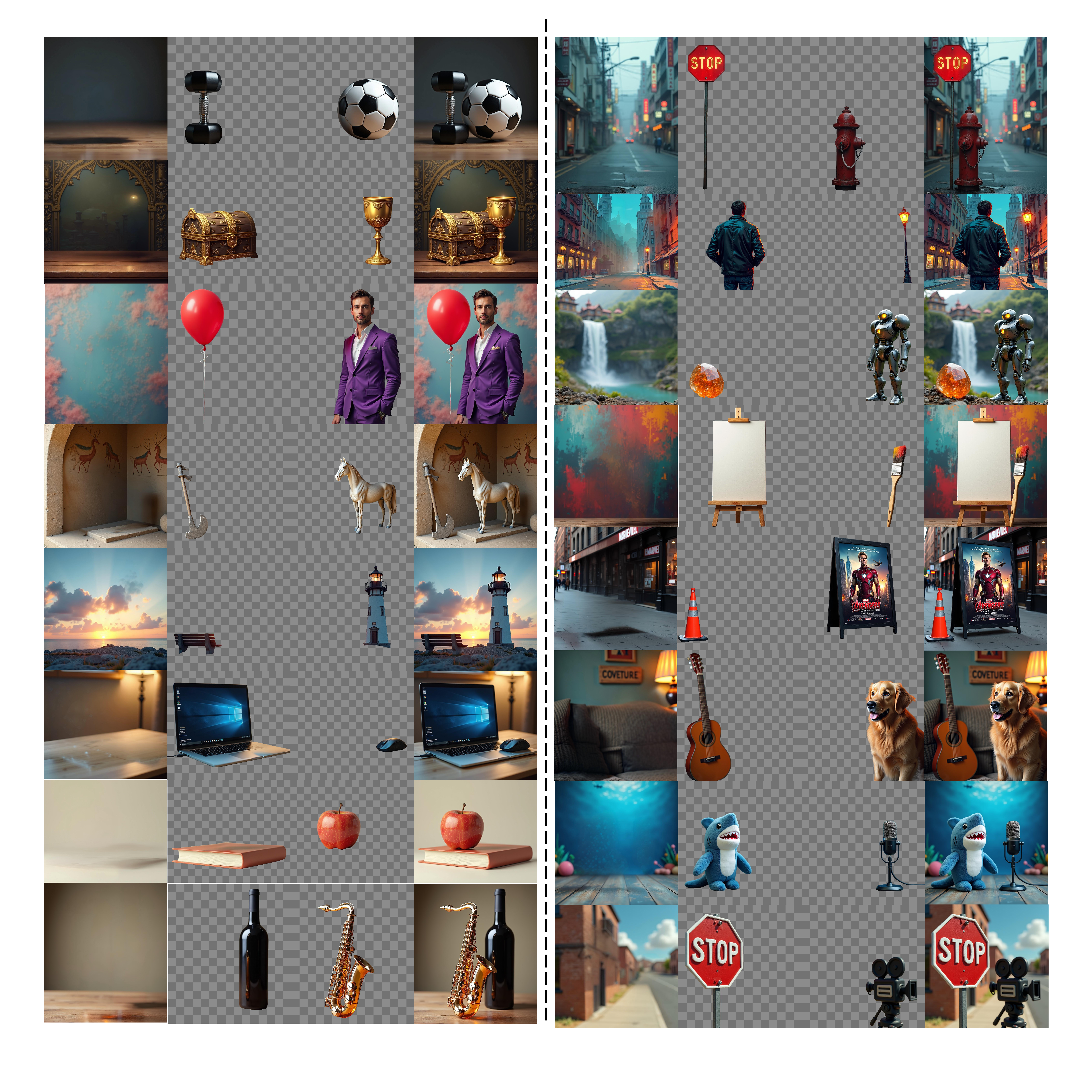}
   \caption{Visulization of three-layer images in our multi-layer dataset. }
   \label{fig_supp: three_dataset}
\end{figure*}

\begin{figure*}[t]
  \centering
   \includegraphics[width=0.9\linewidth]{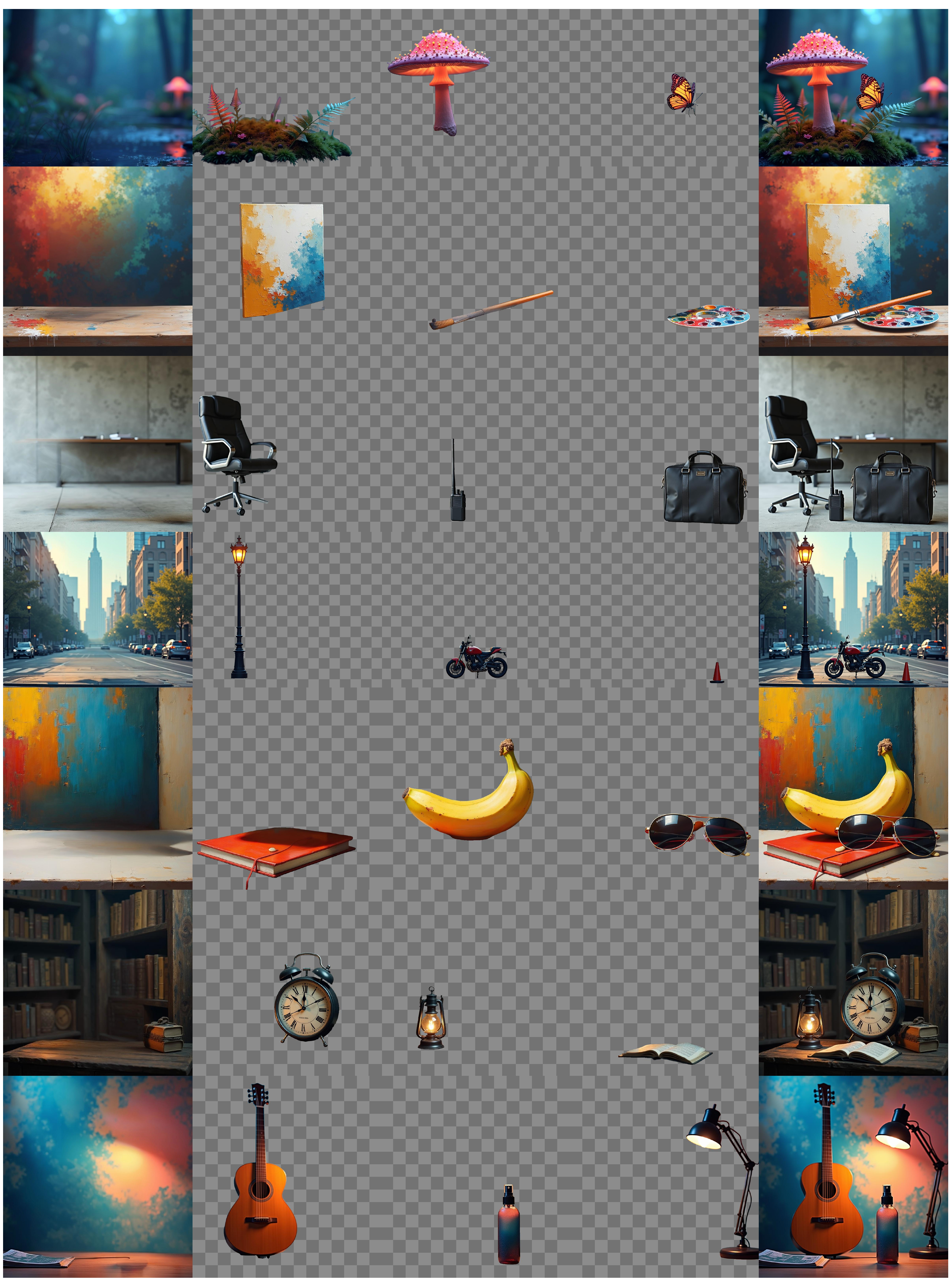}
   \caption{Visulization of four-layer images in our multi-layer dataset. }
   \label{fig_supp: four_dataset}
\end{figure*}

\cref{fig_supp: two_dataset}, \cref{fig_supp: three_dataset} and \cref{fig_supp: four_dataset} showcase examples of multi-layer images generated by our data generation pipeline. With the support of multiple models, our multi-layer dataset achieve high quality and resolution. They also feature logical layer order and precise alpha channels. By leveraging the depth map and sequential inpainting process, our method effectively handles object occlusion. As a result, each layer is nearly complete.

\begin{figure*}[t]
  \centering
   \includegraphics[width=0.7\linewidth]{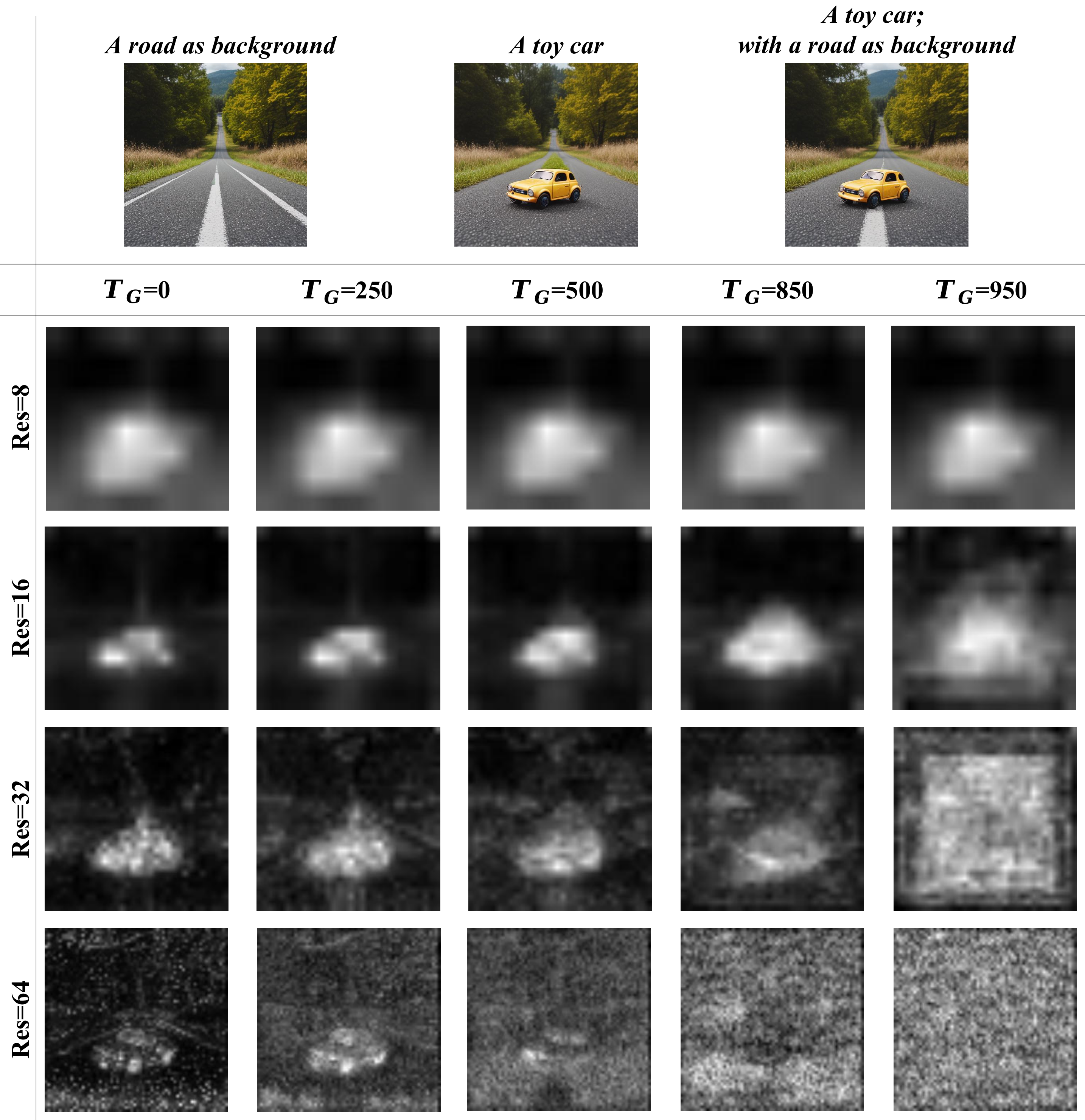}
   \caption{The context map results from global layer extracted at different resolutions and time steps. }
   \label{fig_supp: ab}
\end{figure*}

\section{Implement Details}
During training, we scale and center-crop the images to a size of $512\times512$ as input. The model is initialized with SD1.5 pre-trained weights. Intermediate results with a resolution of 16 are extracted from the four stages of the UNet as the attention maps. Layer-shared self-attention is applied between $T_G=850$ and $T=1000$, while shared self-attention is applied across all steps. All loss weight $\lambda_{noise}, \lambda_l, \lambda_c$ are set to 1. The initial learning rate is set to $2\times10^{-6}$, with a contant rate scheduler applied for gradual decay. Training started with the two-layer data for 60,000 steps, followed by training on the three-layer and four-layer data based on the two-layer model. During inference, we use 50 steps with the DDIM sampling strategy. In the Information Retained Harmonization (IRH) process, latents between $T_H=400$ and $T_H'=600$ are retained, and blending is performed at the latent level at $T_H$.

\section{Quantitative comparison of Bg\&Fg layer}
In the main text, we quantitatively compare the quality of the final composite images. Here, we evaluate the generation quality of background and foreground layers in comparison to LayerDiffusion~\cite{zhang2024transparent}. As shown in Tables \cref{tab_supp:bg}, our method achieves higher aesthetic scores for background generation, particularly excelling in two-layer generation with an improvement of approximately 0.7. Similarly, for foreground generation, our method also outperforms LayerDiffusion, further highlighting the effectiveness in multi-layer generation tasks.

\section{Ablation Study}

\subsection{The Context Map}

We conduct a detailed investigation into the stages and steps $T_G$ for extracting the Context Map from global image. As shown in , among the four stages of the Unet, the clearest context map for foreground object ``toy car'' is extracted at the resolution $res=16$. ther stages primarily capture texture details and image-specific patterns. At $res=16$, the focus is on the layout and general contours of objects.

For different $T_G$ steps, we observe that at $T_G=850$, the context map contains sufficiently clear information. When $T_G$ decreases, the context map becomes sharper. However, this increases the steps of Layer-Shared Self-Attention, introducing more global layer information. As a result, the foreground layer cannot be effectively distinguished from the global layer, leading to layer generation failure. To balance clarity and accuracy, we choose $T_G=850$.

\subsection{Layer-Shared Self-Attention}
LSSA is primarily used to maintain consistency across different image layers, a feature already effective in the original SD15, as shown in \cref{fig: ab_self}. ``Normal Attention" refers to standard self-attention without any inter-layer interaction, where each layer is generated solely based on its respective text prompt. ``Shared Attention" involves layer interaction through concatenation, as described in \cref{eq11}, which brings a certain level of consistency—such as generating similar yellow cars across layers. ``Layer-Shared Attention" further enhances consistency by incorporating global layer information into the foreground layer, as outlined in \cref{eq10}, resulting in better alignment of the size and placement of  the toy car.

\section{Image to Layer}
In the Image to Layer process, we use DDIM inversion~\cite{mokady2023null} to revert the input image into its initial latent. During this process, the input image is treated as a global image and duplicated $k+1$ times to form a layer batch. To ensure clarity, we apple mask attention during inversion to isolate the global layer from other layers, preventing information from other layers from interfering with the global layer during the inversion process. Specifically, in the Layer-Shared Self-Attention process, we first concatenate all the noisy latents $z_t\in\mathbb{R}^{h\times w}$ from different layers:
\begin{equation}
    \tilde{z}_t^c = \mathrm{concat}(\tilde{z}_t^1, \cdots, \tilde{z}_t^{k+1}).
    \label{eq11}
\end{equation}
Next, we generate a mask $M\in\mathbb{R}^{h\times (k+1)w}$ based on $\tilde{z}_t^c$:
\begin{equation}
    M(i,j) =  
    \begin{cases} 
0,  & \mbox{if }j>kw \\
-\infty, & \mbox{otherwise}
\end{cases}
\end{equation}
After applying linear projections, we perform the masked attention operation, formally as:
\begin{equation}
    O_s^i = Softmax(\frac{Q_s^i(K_s^c)^T+M}{\sqrt{d}})V_s^c.
\end{equation}
By applying mask attention to block the influence of other layers on the global layer, we can decompose the input image into multiple layers. As shown in \cref{fig_supp: image2layer}, without mask attention, information from the global layer mixes with other layers during inversion. This often results in foreground information remaining in the background layer.

\section{More Qualitative Results}
\begin{figure*}[t]
  \centering
   \includegraphics[width=0.9\linewidth]{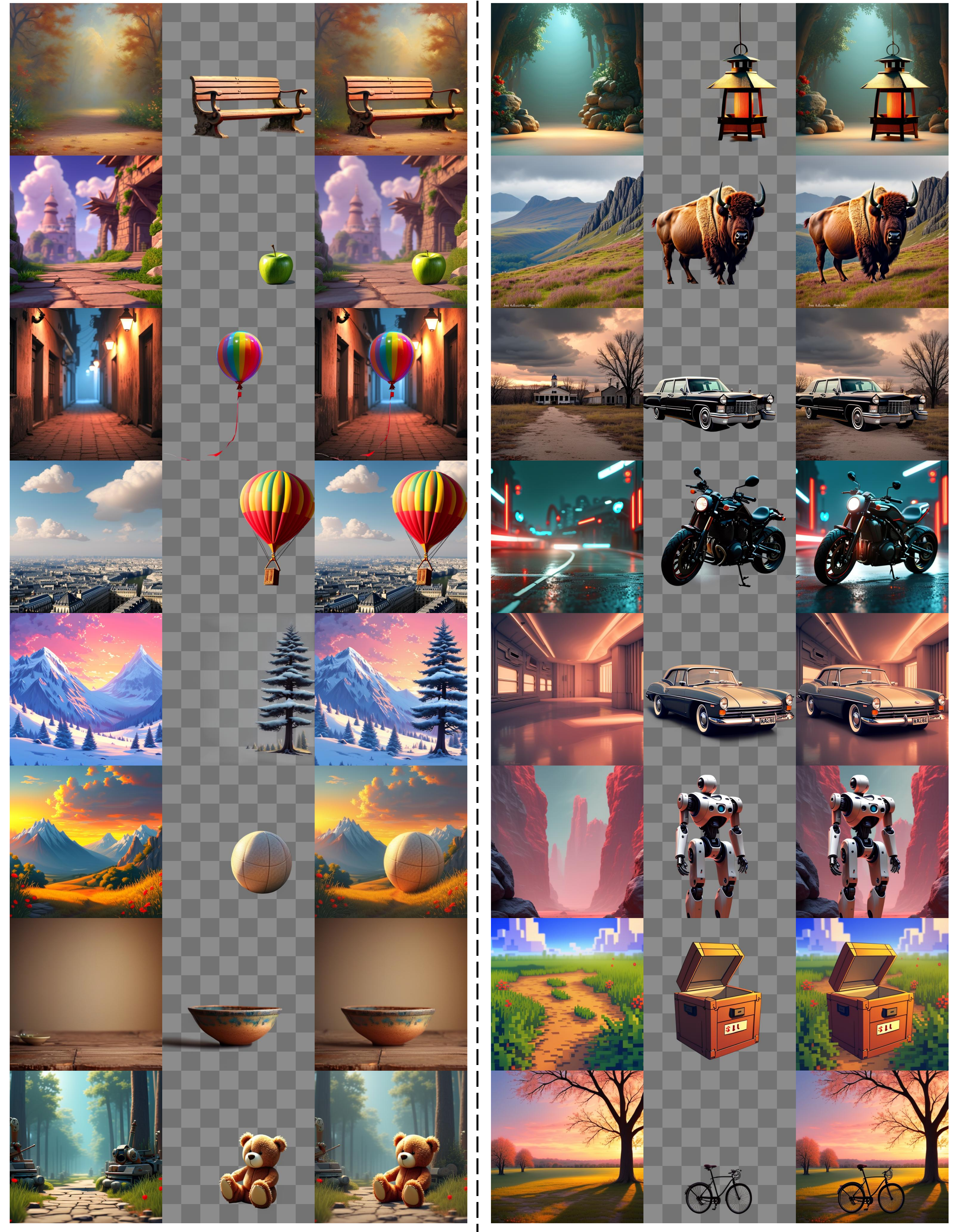}
   \caption{Qualitative Results of two-layer images generated by DreamLayer. }
   \label{fig_supp: two_dl}
\end{figure*}

\begin{figure*}[t]
  \centering
   \includegraphics[width=0.8\linewidth]{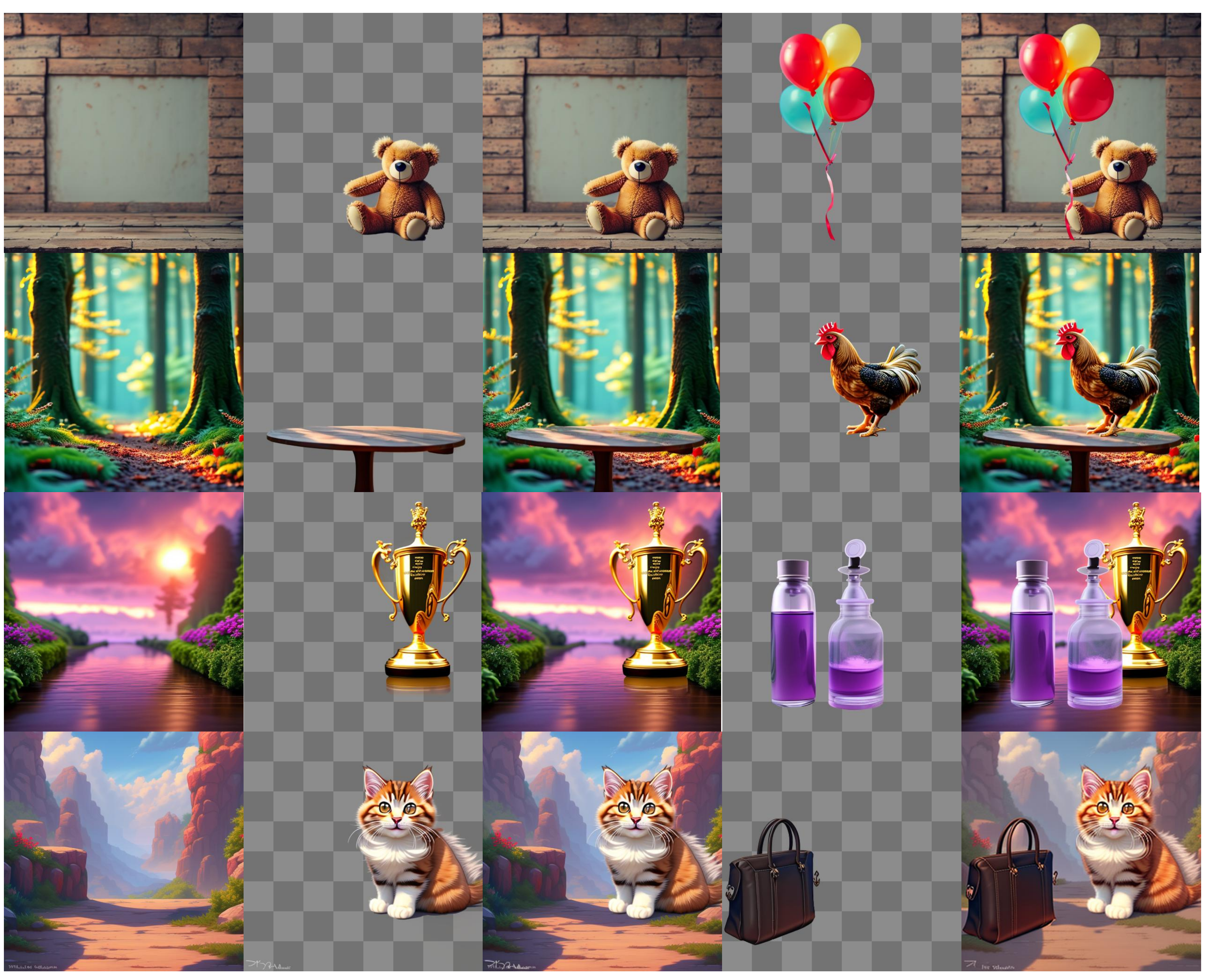}
   \caption{Qualitative Results of three-layer images generated by DreamLayer. }
   \label{fig_supp: three_dl}
\end{figure*}

\begin{figure*}[t]
  \centering
   \includegraphics[width=0.8\linewidth]{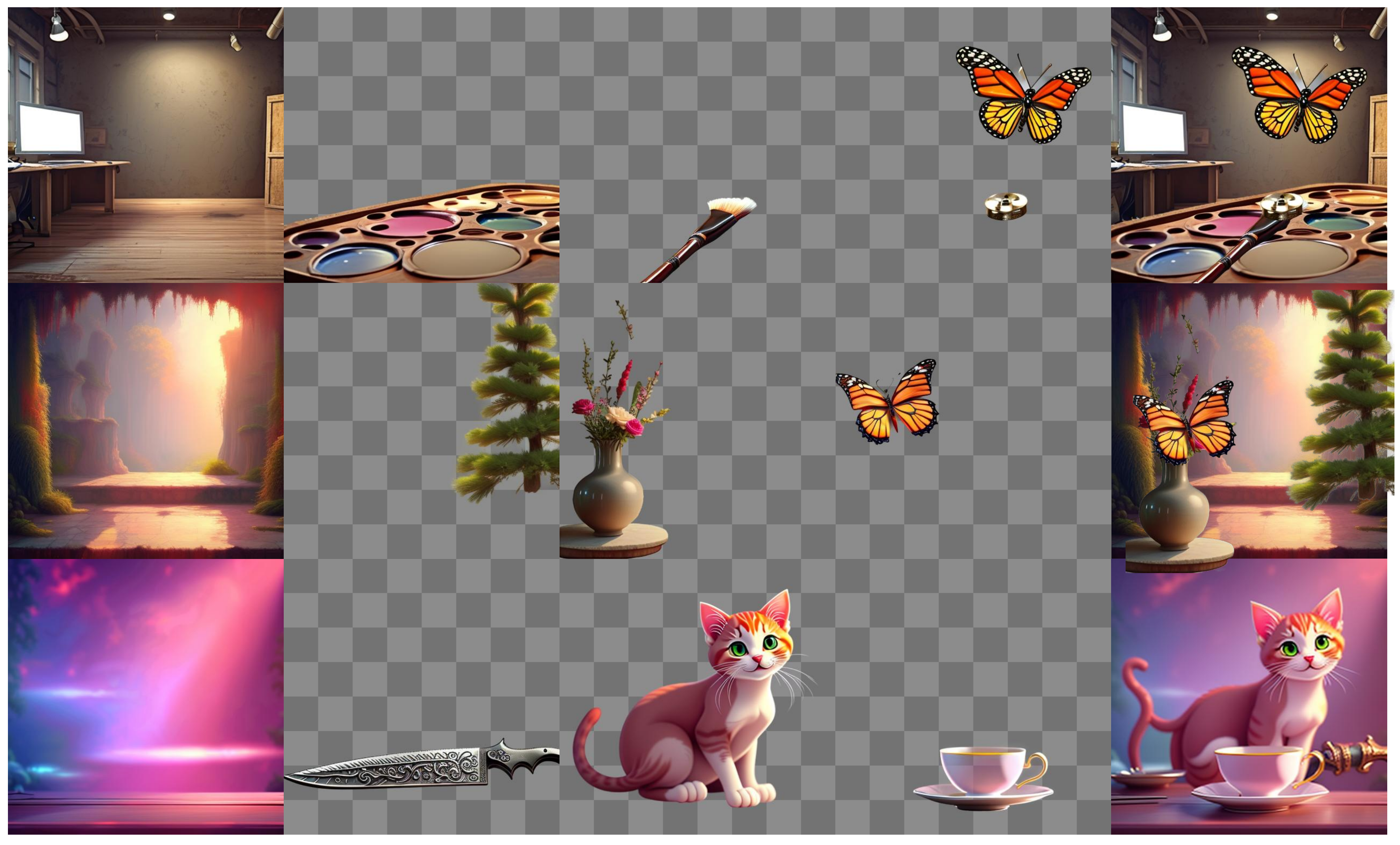}
   \caption{Qualitative Results of four-layer images generated by DreamLayer. }
   \label{fig_supp: four_dl}
\end{figure*}

\cref{fig_supp: two_dl}, \cref{fig_supp: three_dl} and \cref{fig_supp: four_dl}
illustrate the results generated by our DreamLayer on two-layer, three-layer, and four-layer images, respectively.
Under the guidance of the global layer, our generated multi-layer images exhibit well-organized layouts. The foreground objects align more naturally with the background images, resulting in composite images that are more harmonious. These composites also include detailed elements, such as shadows, enhancing their realism.

\end{document}